%% file: main.tex
\documentclass{article} 
\usepackage{iclr2025_delta,times}

\input{math_commands.tex}

\usepackage{hyperref}
\usepackage{url}
\usepackage{cleveref}

\usepackage[ruled,linesnumbered]{algorithm2e}
\usepackage[noend]{algpseudocode}
\usepackage{booktabs}
\usepackage{soul}

\usepackage{float}
\usepackage{graphics}
 \usepackage{graphicx}

\usepackage{caption,setspace}
\usepackage{amsthm}
\usepackage{amssymb}
\usepackage{amsfonts}

\input{math}

\title{
\fontsize{16.8}{16.8}\selectfont
RFMI: Estimating Mutual Information on Rectified Flow for Text-to-Image Alignment
}

\author{%
    $\textrm{\bf Chao Wang}^{1,2}$,
    $\textrm{\bf Giulio Franzese}^{1}$, 
    $\textrm{\bf Alessandro Finamore}^{2}$, 
    $\textrm{\bf Pietro Michiardi}^{1}$
    \\
    $\textrm{EURECOM}^1, \textrm{Huawei Technologies SASU, France}^2$
    \\
        $^1${ \fontsize{8}{8}\selectfont \tt \{chao.wang, giulio.franzeze, pietro.michiardi\}@eurecom.fr}
        \\
        $^2${ \fontsize{8}{8}\selectfont \tt \{wang.chao3, alessandro.finamore\}@huawei.com}
}

\begin{document}

\maketitle

\begin{abstract}
Rectified Flow (RF) models trained with a Flow matching framework have achieved state-of-the-art performance on Text-to-Image (T2I) conditional generation. 
Yet, multiple benchmarks show that synthetic images can still suffer from poor alignment with the prompt, i.e., images show wrong attribute binding, subject positioning, numeracy, etc.
While the literature offers many methods to improve T2I alignment, they all consider only Diffusion Models, and
require auxiliary datasets, scoring models, and linguistic analysis of the prompt.
In this paper we aim to address these gaps. First, we introduce RFMI, a novel 
Mutual Information (MI) estimator 
for RF models that uses the pre-trained model itself for the MI estimation. 
Then, we investigate a self-supervised fine-tuning approach
for T2I alignment based on RFMI that does not require auxiliary information
other than the pre-trained model itself.
Specifically, a fine-tuning set is constructed by selecting 
synthetic images generated from the pre-trained RF model and having
high point-wise MI between images and prompts.
Our experiments on MI estimation benchmarks demonstrate the validity of RFMI, 
and empirical fine-tuning on SD3.5-Medium confirms the 
effectiveness of RFMI for improving T2I alignment while maintaining image quality.
\end{abstract}


\input{sections/intro_new}

\input{sections/preliminaries_v2}
\section{RFMI: Estimating Mutual Information with Rectified Flows}\label{sec:mi_estimator}



\input{sections/method_v2}

\input{sections/mitune}

\input{sections/experiments}

\input{sections/conclusion}

\bibliography{iclr2025_delta}
\bibliographystyle{iclr2025_delta}

\clearpage
\appendix
\section{Appendix}

\input{sections/proofs_v1}

\input{sections/experiments_hyperp}

\end{document}

%% file: math_commands.tex
\usepackage{amsmath,amsfonts,bm}

\def\eqref#1{equation~\ref{#1}}

\def\1{\bm{1}}

\DeclareMathAlphabet{\mathsfit}{\encodingdefault}{\sfdefault}{m}{sl}
\SetMathAlphabet{\mathsfit}{bold}{\encodingdefault}{\sfdefault}{bx}{n}



%% file: math.tex
\newcommand{\mathbold}[1]{{\boldsymbol{{#1}}}}

\newcommand{\nestedmathbold}[1]{{\mathbold{#1}}}

\newcommand{\mbu}{\nestedmathbold{u}}

\newcommand{\mby}{\nestedmathbold{y}}
\newcommand{\mbz}{\nestedmathbold{z}}

\newcommand{\mbI}{\nestedmathbold{I}}

\newcommand{\mbtheta}{\nestedmathbold{\theta}}

\newcommand{\mt}{m_t}
\newcommand{\ft}{\frac{\dd \log \mt}{\dd t}}

\usepackage[english]{babel}
\usepackage{lmodern}
\usepackage{amsthm}

\theoremstyle{proposition}
\newtheorem{proposition}{Proposition}[section]

\theoremstyle{definition}

%% file: sections/intro_new.tex
\section{Introduction}

Text-to-Image (T2I) generative models have 
reached an incredible popularity thanks to their high-quality image synthesis,
ease of use, and integration across a variety of end-users services 
(e.g., image editing software, chat bots, smartphones apps, websites).
T2I models are trained using large-scale datasets~\citep{laion} to
generate images \emph{semantically aligned} with a user text input. 
Yet, recent 
benchmarks~\citep{huang2023t2icompbench,wu2024conceptmixcompositionalimagegeneration}
show that even Rectified Flow (RF) models, e.g., Stable Diffusion 3~\citep{SD3} and FLUX~\citep{flux2023},
despite achieving a new state of the art, 
still suffer from a variety of alignment issues
(subjects in the images might be missing, or have the wrong attributes, such as numeracy, positioning, etc).

Many works in the literature propose methods to mitigate alignment issues
at either inference-time -- e.g., steering
the generation guided by auxiliary text information or map-based objectives~\citep{chefer2023attendandexcite,feng2023Structured,shen2024SCFG} --
or using model fine-tuning -- e.g., updating the pre-trained model weights 
via supervised learning~\citep{krojer2023HNitm}
or reinforcement learning~\citep{fan2023dpok,huang2023t2icompbench}
to address a specific task.
Despite their merits, available alignment methods
require \emph{complementary data} such as 
linguistic analysis of prompts 
-- e.g., A\&E~\citep{chefer2023attendandexcite} steers U-Net cross-attention units 
by knowing which are the relevant tokens the prompt that
need to appears in the image, auxiliary models -- e.g., DPOK~\citep{fan2023dpok}
uses a reward function based on a complementary model trained to
capture human judgment of alignment and aesthetics,
or datasets -- e.g., HNITM~\citep{krojer2023HNitm} relies on a contrastive learning
approach creating negative prompt examples from a set of target prompts. In other words,
these methods shift the alignment problem from the model to the users.







Differently from existing literature, 
in this work we argue for avoiding auxiliary input
in favor of a \emph{self-supervised approach} where the pre-trained model
is used to compute a score signaling if the generated images align with the text prompt.
In particular, we aim to use a pre-trained RF model as a neural estimator for the
Mutual Information (MI) between the generated image and the prompt. 
In turn, this raises two research questions:
how to estimate MI using an RF model? and how to use the MI estimates
to improve T2I alignment? 

While multiple MI neural estimators have been proposed,
to the best of our knowledge no previous work considers RF models.
For instance, 
discriminative approaches~\citep{mcallester2020formallimitationsmeasurementmutual}
focus on directly estimating the ratio between joint distribution and product of marginals
but 
the sample average's variance scales exponentially with the ground-truth MI.
Considering methods closer to our scope, generative approaches based on estimating 
the two marginal densities separately
with 
generative models like  
normalizing flows~\citep{FlowbasedVariationalMI, dinh2017densityestimationusingreal} 
are more scalable but
face estimation accuracy challenges on high-dimensional or complex data according to benchmark testing on synthetic distributions~\citep{czyż2023normalevaluationmutualinformation}.
However, none of these methods directly applies to RF models.

More important, the literature on T2I alignment primarily focuses on Diffusion Models (DMs) such as SD2,
while RF-related literature only tangentially considers the problem.
For instance, \citep{li2024omniflowanytoanygenerationmultimodal} extends SD3 with 
more modalities, \citep{dalva2024fluxspacedisentangledsemanticediting} enhances
FLUX with a linear and fine-grained editing scheme of models' attention output,
\citep{liu2024instaflow} proposes a novel text-conditioned pipeline
to turn Stable Diffusion (SD) into an ultra-fast one-step model -- while all these
works present CLIP score evaluations, they neither focus nor they are designed to
address T2I alignment. At the same time, the intrinsic different nature of the RF models architecture
(e.g., SD3 replaces the U-Net of SD2 with a DiT architecture and also add an extra text transformer)
calls for redesigning some of the mechanics of DM-based T2I aligment methods.

In summary, in this work:  
\begin{itemize}
\item We introduce RFMI, an RF-based point-wise MI estimator
 (Section \ref{sec:mi_estimator}) leveraging the relation between the score of the density $\nabla \log p_t$ and  the velocity field $u_t$.

\item We design a self-supervised fine-tuning approach, called RFMI FT (Section \ref{sec:mitune}), that uses a small number of fine-tuning samples to improve the pre-trained T2I model alignment with no inference-time overhead, nor auxiliary models other than the generative model itself.

\item We demonstrate the validity of our MI estimator considering both a synthetic benchmark involving various challenging data distributions (Section \ref{sec:exp_1}) where the true MI is known, as well as a T2I benchmark~\citep{huang2023t2icompbench} (Section~\ref{sec:exp_2}).

\end{itemize}

%% file: sections/preliminaries_v2.tex
\section{Preliminaries}\label{sec:preliminaries}

This work relies on recent advances in generative modeling~\citep{SD3, flux2023} as a building block to design an estimator for the mutual information between two random variables. Here, we give the necessary background to develop our methodology, that revolves around flow models~\citep{chen2019neuralordinarydifferentialequations} and flow matching framework~\citep{lipman2023flowmatchinggenerativemodeling}.

Let $x\in\mathbb{R}^d$ denote a data point in the $d$-dimensional Euclidean space 
associated with the standard Euclidean inner product,
and $X \in \mathbb{R}^d$ a Random Variable (RV)
with continuous Probability Density Function (PDF) $p_X: \mathbb{R}^d \rightarrow \mathbb{R}_{\geq 0}$, where $\int_{\mathbb{R}^d} p_X(x) \mathrm{d} x=1$. 
We use the notation $X \sim p_X$ to indicate that $X$ is distributed according to $p_X$.

The key concepts we consider within the framework of flow matching are: 







\begin{math}
\begin{array}{lll}
\textrm{1. flow}: 
& \textrm{a time-dependent } C^r\left([0,1] \times \mathbb{R}^d, \mathbb{R}^d\right) \textrm{ mapping }  
\psi \colon 
(t,x)  \mapsto \psi_t(x)\\
\textrm{
2. velocity field}: 
& \textrm{a time-dependent } C^r\left([0,1] \times \mathbb{R}^d, \mathbb{R}^d\right) \textrm{ mapping } 
u \colon 
(t,x)  \mapsto u_t(x)\\
\textrm{3. probability path}: 
& \textrm{a time-dependent PDF }   \left(p_t\right)_{0 \leq t \leq 1} \\
\end{array}
\end{math}

Given 
a 
source distribution $p$ 
-- e.g., standard Gaussian distribution $\mathcal{N}(0, I)$,
and 
the 
data target distribution $q$,
the goal of generative flow modeling is to build a flow that
transforms 
$X_0 \sim p$ into 
$X_1:=\psi_1\left(X_0\right)$ such that $X_1 \sim q$.

Considering 
an arbitrary probability path $p_t$,
$u_t$ is said to \textit{generate} $p_t$ if 
its flow $\psi_t$ satisfies
$
X_t:=\psi_t\left(X_0\right) \sim p_t
$
for $t \in[0,1)$, 
$X_0 \sim p_0$, since there is an equivalence between flows and velocity fields
derived from 
the ordinary differential equation (ODE)
\begin{equation}
  \begin{cases}
\frac{\mathrm{d}}{\mathrm{d} t} \psi_t(x)=u_t\left(\psi_t(x)\right)  \quad  \text{(flow ODE)} 
\\
\psi_0(x)=x   \quad\quad\quad\quad\quad\:\:\,   \text{(flow initial conditions)}.\\
  \end{cases}
\end{equation}
One practical way 
to verify if a vector field $u_t$ generates a probability path $p_t$
is 
to verify if the pair $\left(u_t, p_t\right)$ satisfies 
the Continuity Equation for $t \in[0,1)$
\begin{equation}\label{eq:continuity}
\frac{\mathrm{d}}{\mathrm{~d} t} p_t(x)+\operatorname{div}\left(p_t u_t\right)(x)=0.
\end{equation}
Given a prescribed probability path $p_t$ satisfying the boundary conditions $p_0=p$ and $p_1=q$, 
the goal of Flow Matching (FM) is to learn a parametric velocity field $u_t^\theta$ 
that matches the ground truth velocity field $u_t$ known to generate the desired probability path $p_t$.
This goal is realized by minimizing the
regression loss: $\mathcal{L}_{\mathrm{FM}}(\theta)=\mathbb{E}_{X_t \sim p_t} 
\| u_t^\theta\left(X_t\right) - u_t\left(X_t\right) \|^2
$.

In practice, the ground truth marginal velocity field
$u_t$ is not tractable, as it requires marginalizing over the entire training set -- i.e., $u_t(x)=\int u_t\left(x \mid x_1\right) p_{1 \mid t}\left(x_1 \mid x\right) \mathrm{d} x_1$.
Instead, we consider the Conditional Flow Matching (CFM) loss: 
$
\mathcal{L}_{\mathrm{CFM}}(\theta)=\mathbb{E}_{t, X_1, X_t \sim 
p_{t \mid 1} 
(\cdot \mid X_1)} \left\|u_t^\theta\left(X_t\right)-u_t\left(X_t \mid X_1\right)\right\|^2
$, 
where the ground truth conditional velocity field
$u_t(\cdot \mid x_1)$
is tractable, as it only depends on a single data sample $X_1=x_1$.
The two losses are equivalent for learning purposes: 
since their gradients coincide, 
the minimizer of CFM loss is the marginal velocity $u_t(x)$.



Training using CFM loss requires
($i$) designing $p_{t \mid 1}\left(\cdot \mid x_1\right)$ yielding a marginal probability path $p_t$ satisfying the boundary conditions
and then
($ii$) finding $u_t\left(\cdot \mid x_1\right)$ generating $p_{t \mid 1}\left(\cdot \mid x_1\right)$.
These two tasks could be reduced to defining 
a $C^2\left([0,1) \times \mathbb{R}^d, \mathbb{R}^d\right)$ conditional flow $\psi:[0,1) \times \mathbb{R}^d \rightarrow \mathbb{R}^d$ satisfying
$
\psi_t\left(x \mid x_1\right)= \begin{cases}x & t=0 \\ x_1 & t=1\end{cases}
$.

A popular objective is to minimize a bound of the Kinetic Energy, which results in the flow
\begin{equation}
    \psi_t\left(x \mid x_1\right)=t x_1+(1-t) x.
\label{eq:kinetic}
\end{equation}
When $x$ is a data sample $x_0$ of $X_0 \sim p$, 
\cref{eq:kinetic} becomes 
$\psi_t\left(x_0 \mid x_1\right)=t x_1+(1-t) x_0$, 
the conditional velocity field reduces to 
$u_t\left(x_t \mid x_1\right)=x_1-x_0$ for all $t \in [0,1)$, 
and the CFM loss simplifies to
\begin{equation}
    \mathcal{L}_{\mathrm{CFM}}(\theta)=
\mathbb{E}_{t, 
\left(X_0, X_1\right) \sim \pi_{0,1}
}\left\|u_t^\theta\left(X_t\right)-\left(X_1-X_0\right)\right\|^2, 
\end{equation}
where 
$\pi_{0,1}$ denotes the joint 
distribution 
known as the source-target coupling.
The conditional probability path  $p_{t \mid 1}$
of this linear conditional flow,  
defined using the push-forward formula, 
satisfies the boundary conditions; furthermore, it 
is a particular case of Gaussian paths
$p_{t \mid 1}\left( \cdot \mid x_1\right)=\mathcal{N}\left(  \cdot \mid b_t x_1,  a_t^2 I\right)$ with $b_t=t$ and $a_t=1-t$, if the source distribution $p$ is the standard Gaussian. 


Since a property of this linear conditional flow is that the Kinetic Energy of the marginal velocity $u_t(x)$ is not bigger than 
the Kinetic Energy of the original  coupling $\pi_{0,1}$ used to train the model,
\cite{liu2022flowstraightfastlearning} denote  
$u_t^\theta$ optimized with CFM loss as Rectified Flow~(RF).

As we are interested in conditional generation,
we 
define the guidance RV as $Y \sim p_Y$, with data samples $y \in \mathcal{Y} \subset  \mathbb{R}^k$.
Given access to labeled target samples $\left(x_1, y\right)$,
the goal of conditional FM is to 
train the parameters $\theta$ of a single velocity field 
$u_t^\theta : \mathbb{R}^d \times \mathbb{R}^k \rightarrow \mathbb{R}^d$ 
to match the ground truth guided velocity field $u_t(\cdot \mid y)$ known to generate the desired guided probability path $p_{t \mid Y}(\cdot \mid y)$ 
satisfying the boundary conditions 
$p_{t=0  \mid Y}(\cdot \mid y) = p\left(\cdot\right)$ 
and 
$p_{t=1 \mid Y}(\cdot \mid y) = q\left(\cdot \mid y\right)$ 
, for all values of $y$.
The guided version of CFM loss is
$\mathcal{L}_{\mathrm{CFM}}(\theta)=\mathbb{E}_{t,\left(X_0, X_1, Y\right) \sim \pi_{0,1, Y}} 
\left\|
u_t^\theta\left(X_t \mid Y\right) - 
\left(X_1-X_0\right)
\right\|^2
 $. 
Furthermore, if the model is trained with Gaussian paths, 
the Classifier-Free Guidance (CFG) technique can be applied at inference to enhance the sample quality, 
 for which
 during training $y$ will be masked as null-condition $\emptyset$ with probability $p_{\text {uncond}}$,
 in order to train $u_t^\theta(\cdot \mid \emptyset)$ to approximate the 
 unconditional velocity field $u_t$
 generating
 the unconditional probability path $p_t$. 

Finally, an essential result known as the Instantaneous Change of Variables~\citep{chen2019neuralordinarydifferentialequations} is required for our work. 
Given the Continuity Equation, the differential equation governing the evolution of the log-probability density is:
\begin{equation}\label{eq:Instantaneous}
\frac{\mathrm{d}}{\mathrm{~d} t} \log p_t\left(\psi_t(x)\right)=-\operatorname{div}\left(u_t\right)\left(\psi_t(x)\right). 
\end{equation}

%% file: sections/method_v2.tex








Recall that the MI between two random variables $X \sim p_X$ and $Y \sim p_Y$ can be defined as the Kullback–Leibler (KL) divergence between the joint 
distribution and the product of marginals: $I(X ; Y)=D_{\mathrm{KL}}\left(p_{(X, Y)} \| p_X p_Y\right)$. 
Furthermore, let $p
_{X \mid Y}
(\cdot | y)$ be the conditional 
{distribution} of $X$ given $Y=y$. Using the identity 
$p_{(X, Y)}(x, y) = 
p_{X \mid Y}(x \mid y) p_{Y}(y)$, it clearly holds that  $\mathrm{I}(X ; Y)=\mathbb{E}_Y\left[D_{\mathrm{KL}}\left(p_{X \mid Y} \| p_X\right)\right]$, which indicates that the more the distributions $p_{X \mid Y}$ and $p_X$ differ on average, the greater the information gain.

In this section we introduce RFMI, a new method to estimate 
the MI between $X$ and the guidance signal $Y$ leveraging conditional RF models~\citep{SD3, flux2023}. To keep the notation concise and aligned with the notions in \Cref{sec:preliminaries}, we will refer to
$p
_{X}
$
as 
$q$, 
and 
$p
_{X \mid Y}
(\cdot | y)$
as 
$q
(\cdot | y)$.

Consider 
the case of 
linear conditional flow $\psi_t\left( x \mid x_1\right)=t x_1+(1-t) x$ with $x$ being samples of Gaussian prior 
$X_0 \sim p =\mathcal{N}(0, I)$.
This flow's conditional velocity field $u_t(\cdot \mid x_1)$ 
generates a Gaussian path $p_{t \mid 1}\left(\cdot \mid x_1\right)$ 
satisfying $p_{0 \mid 1}\left(\cdot \mid x_1\right)=p$ and $p_{1 \mid 1}\left(\cdot \mid x_1\right)=\delta_{x_1}(\cdot)$.
The conditional pair $\left(u_t(\cdot \mid x_1), p_{t \mid 1}\left(\cdot \mid x_1\right)\right)$ does not depend from the guidance variable $Y$, 
while its marginal counterpart does, as
$\left(
u_t(\cdot  \mid y)=\int u_t\left(\cdot \mid x_1\right) p_{1 \mid t, Y}\left(x_1 \mid \cdot , y\right) \mathrm{d} x_1
,\, 
p_{t \mid Y}(\cdot  \mid y)=\int p_{t \mid 1}\left(\cdot  \mid x_1\right) q\left(x_1 \mid y\right) \mathrm{d} x_1
\right)$.
As a consequence, by applying the Marginalization trick,
the guided velocity field $u_t(\cdot  \mid y)$ generates the guided probability path $p_{t \mid Y}(\cdot \mid y)$, 
$p_{t \mid Y}(\cdot  \mid y)$ satisfies
$p_{0  \mid Y}(\cdot \mid y) = p\left(\cdot\right)$ 
and 
$p_{1 \mid Y}(\cdot \mid y) = q\left(\cdot \mid y\right)$. 
Note that 
when $y=\emptyset \in \{\emptyset\}$
the marginal case is reduced to 
unconditional generation: 
$u_{t}(\cdot \mid \emptyset)=u_t$,
$p_{t \mid  \{\emptyset\}}(\cdot \mid \emptyset)=p_t$,
and $q\left(\cdot \mid \emptyset\right) = q$.
Overall, we express MI using the guided and the unconditional marginal probability paths at the endpoint $t=1$ as

\begin{equation}\label{eq:I1}
\begin{aligned}
\mathrm{I}(X ; Y)
&=\mathbb{E}_Y\left[D_{\mathrm{KL}}\left(p_{X \mid Y} \| p_X\right)\right] \\
&=
\mathbb{E}_Y\left[
    \int 
    q(x \mid Y) 
    \log \left(\frac{
    q(x \mid Y ) 
    }{
    q(x)
    }\right) \mathrm{d} x
\right] 
\\
&=
\mathbb{E}_Y\left[
    \int 
    p_{1 \mid Y}(x_1 \mid Y)
    \log \left(\frac{
    p_{1 \mid Y}(x_1 \mid Y)
    }{
    p_{1}(x_1) 
    }\right) \mathrm{d} x_1
\right].
\end{aligned}
\end{equation}

In practice, 
we train a single conditional RF neural network $u_t^\theta(x \mid y)$, using the CFM loss, 
for all values $y \in\{\mathcal{Y}, \emptyset\}$. 
Since the minimizer of CFM loss is $u_t(\cdot  \mid y)$, 
$u_t^\theta(x \mid y)$ is a valid approximation of $u_t(\cdot  \mid y)$.

Next, 
we develop an expression of MI using $u_t(\cdot  \mid y)$ 
and $u_t$,
and 
use the 
conditional RF model 
to estimate the MI between $X$ and $Y$.
To do so, we first express the score functions associated to the  marginal probability paths 
using the marginal velocity fields. 
These two terms  
are related according to the following 

\begin{proposition}[Relation between velocity field and score function] \label{prop:rho-v}
{
    For Gaussian paths $p_{t \mid 1}\left( \cdot \mid x_1\right)=\mathcal{N}\left(  \cdot \mid b_t x_1,  a_t^2 I\right)$, 
    the relation between 
    the conditional velocity field $u_t(\cdot  \mid x_1)$ 
    and 
    the \textbf{score function} $\nabla \log p_{t \mid 1}\left( \cdot \mid x_1\right)$
    of the conditional probability path $p_{t \mid 1}(\cdot \mid x_1)$
    is derived as :
    \begin{equation}\label{eq:score_cond}
       u_t(x  \mid x_1) =\frac{\dot{b}_t}{b_t} x +\left(\dot{b}_t a_t-b_t \dot{a}_t\right) \frac{a_t}{b_t} 
       \nabla \log p_{t \mid 1}\left( x \mid x_1\right)
       .
    \end{equation}
    This relation also holds for their marginal counterpart, both in the guided case and in the unconditional case:
    \begin{equation}\label{eq:score}
     \begin{cases}
     \begin{aligned}
          u_t(x  \mid y) &=\frac{\dot{b}_t}{b_t} x +\left(\dot{b}_t a_t-b_t \dot{a}_t\right) \frac{a_t}{b_t} 
       \nabla \log p_{t \mid Y}(x \mid y) \\
       u_t(x ) &=\frac{\dot{b}_t}{b_t} x +\left(\dot{b}_t a_t-b_t \dot{a}_t\right) \frac{a_t}{b_t} 
       \nabla \log p_{t}(x)
       .   
     \end{aligned}
       \end{cases}
    \end{equation}
See proof in Eq. (F.4)~\citep{albergo2023buildingnormalizingflowsstochastic}, Eq. (7)~\citep{zheng2023guidedflowsgenerativemodeling}.
}
\end{proposition}

In particular, for 
linear conditional flow with Gaussian prior, 
i.e. $b_t=t$, $a_t=1-t$, 
we have $\dot{b}_t=1$, $\dot{a_t}=-1$, and \Cref{eq:score} becomes 
 \begin{equation}\label{eq:score2}
     \begin{cases}
      \begin{aligned}
    \nabla \log p_{t \mid Y}(x \mid y) &=
    \frac{t u_t(x \mid y)- x }{1-t} \\
   \nabla \log p_{t}(x) &=
    \frac{t u_t(x)- x }{1-t}
       .
       \end{aligned}
       \end{cases}
    \end{equation}
We note that \Cref{eq:score2} is only defined for $t\in[0,1)$. 
As $t\rightarrow1$, by taking the limit of \Cref{eq:score2} using 
l’Hopital’s rule, the limit of score function is (proof in Appendix \ref{sec:proof_lemma1}):
\begin{equation}\label{eq:score3}
  \begin{cases}
      \begin{aligned}
   \lim_{t \rightarrow 1}  \nabla \log p_{t \mid Y}(x \mid y) &
  = 
  \lim_{t \rightarrow 1} 
   -
    \partial_t u_t(x \mid y )
    \\
    \lim_{t \rightarrow 1}   \nabla \log p_{t}(x) &=  
     \lim_{t \rightarrow 1} 
     -
    \partial_t u_t(x).
       \end{aligned}
       \end{cases}
\end{equation}



Given the guided and unconditional ground truth marginal velocity fields $u_t(\cdot|y)$ and $u_t$, 
it is possible to show that MI can be computed exactly,
as done in the following

\begin{proposition}[MI computation]\label{prop:micomp}
{
Given a linear conditional flow with Gaussian prior, the MI between the target data $X$ and the guidance signal $Y$ is given by 
\begin{equation}\label{eq:mi}
\begin{aligned}
\mathrm{I}(X ; Y)
&=
\mathbb{E}_Y\left[
    \int 
    p_{1 \mid Y}(x_1 \mid Y)
    \log \left(\frac{
    p_{1 \mid Y}(x_1 \mid Y)
    }{
    p_{1}(x_1) 
    }\right) \mathrm{d} x_1
\right] 
\\
&=
\mathbb{E}_Y\left[
   \int_0^1 
        \mathbb{E}_{X_t \mid Y}
        \left[   
            \frac{t}{1-t}
            u_t(X_t|Y)  \cdot
                \left( u_t(X_t|Y) - u_t(X_t) 
                \right)
        \right]
    dt
\right]. 
\end{aligned}
\end{equation}

}
\end{proposition}
This can be proven leveraging \Cref{eq:continuity}, \Cref{eq:Instantaneous}, and the result of \Cref{prop:rho-v}. The full proof of \Cref{prop:micomp} is given in Appendix \ref{sec:proof_prop2}.

Similarly, it is easy to show that, given an individual guidance sample $Y=y$ , 
it is possible to use~\Cref{eq:mi} 
to compute the \textbf{point-wise MI} as
\begin{equation}
\label{eq:pointwise_mi}
    \mathrm{I}(X ; y) 
=
   \int_0^1 
        \mathbb{E}_{X_t \mid Y=y}
        \left[   
            \frac{t}{1-t}
            u_t(X_t|Y=y)  \cdot
                \left( u_t(X_t|Y=y) - u_t(X_t) 
                \right)
        \right]
    dt
.
\end{equation}


The integral in \Cref{eq:mi} can be estimated by uniform sampling $t \sim \mathcal{U}(0, 1)$.
However, in practice, since the denominator $(1-t) \rightarrow 0$ as $t \rightarrow 1$, this estimator has unbounded variance. To reduce variance, since the argument of the integral has constant magnitude on average, it would be tempting to use importance sampling where $t \sim f(t) \propto \frac{t}{1-t}$. This ratio, however, is hard to normalize (as it integrates to $\infty$). As an alternative, we consider the following un-normalized density $\tilde{f}_\epsilon$ proportional to such ratio for most 
of its support, and then truncated to a constant for large $t$:
\begin{equation}\label{eq:truncated_pdf}
    \tilde{f}_\epsilon(t)= 
    \begin{cases}
        \frac{t}{1-t} & t \in\left[0, t_\epsilon\right) \\ 
        \frac{t_\epsilon}{1-t_\epsilon} & t \in\left[t_\epsilon, 1\right]
    \end{cases}
\end{equation}
To implement such non-uniform sampling in practice, we use the inverse transform sampling method, with the inverse Cumulative Distribution Function (CDF) described in 

\begin{proposition}
[Non-uniform sampling for 
importance sampling]
\label{sec:lemma2}
The inverse CDF of a PDF proportional to truncated $\frac{t}{1-t}$ is 
    \begin{equation}\label{eq:sampling}
        F_\epsilon^{-1}(u)= 
        \begin{cases}
            1 + W(-e^{-Zu-1}) & u \in\left[0, \frac{-\ln (1-t_\epsilon)-t_\epsilon }{Z}\right), 
        \\
        \rule{0pt}{1.5em}
            1 + \frac{1-t_\epsilon}{t_\epsilon} \left[\ln (1-t_\epsilon) + Zu\right] & u \in\left[\frac{-\ln (1-t_\epsilon)-t_\epsilon }{Z}, 1\right],
        \end{cases}
    \end{equation}
in which $W$ is the Lambert's $W$-function\footnote{\fontsize{6}{6}\selectfont\url{https://docs.scipy.org/doc/scipy/reference/generated/scipy.special.lambertw.html}}, and the normalizing constant is $Z=-\ln (1-t_\epsilon)$.    
\end{proposition}
We show the proof 
of 
\Cref{sec:lemma2} 
in Appendix \ref{sec:proof_lemma2}.

Finally, given the parametric approximations of marginal velocity fields through minimization of CFM loss, 
and the result in \Cref{prop:micomp}, now we are able to propose an MI estimator defined as
\begin{equation}\label{eq:mi-estimator}
\mathrm{I}(X ; Y)
\approx
\mathbb{E}_Y\left[
   \int_0^1 
        \mathbb{E}_{X_t \mid Y}
        \left[   
            \frac{t}{1-t}
            u_t^\theta(X_t|Y)  \cdot
                \left( u_t^\theta(X_t|Y) - u_t^\theta(X_t) 
                \right)
        \right]
    dt
\right] 
\end{equation}




For the estimation of point-wise MI between generated image and guidance prompt, in practice we found that
using the velocity field calibrated by CFG as $u_t^\theta(X_t|Y)$ in \Cref{eq:mi-estimator} 
leads to better performance than 
using the vanilla guided output given by the model. 
Furthermore, for generating images at high resolution, 
to reduce training's computational cost and to speed up inference,
it is common to apply RF on a lower dimensional manifold (e.g. the compressed latent space of a pretrained Variational Autoencoder). 
It is easy to show that the MI between images $X$ and prompts $Y$ equals to that between images' latents $Z$ and prompts $Y$, i.e.
$I(X; Y) = I(Z; Y)$.

An important property of our estimator is that it is neither an upper nor a lower bound of the true MI, since the difference between the ground truth velocity fields and their parametric approximation can be positive or negative. This property frees our
estimation method from the pessimistic results of \cite{mcallester2020formallimitationsmeasurementmutual}.






%% file: sections/mitune.tex
\input{algos/mitune}
\vspace{-0.2cm}
\section{Improving Text-Image Alignment with MI-guided Self-Supervised Fine-tuning}
\label{sec:mitune}
\vspace{-0.2cm}



Given a pre-trained RF model -- e.g., Stable Diffusion 3~\citep{SD3} or Flux.1~\citep{flux2023}, we leverage RFMI to improve the model's alignment via fine-tuning. Specifically, our \emph{self-supervised} approach \textbf{relies on the pre-trained model} to create a synthetic fine-tuning set of prompt-image pairs with a high degree of alignment selected using the \textbf{point-wise MI estimates obtained from the pre-trained model itself}.

As described in Algorithm \ref{algo:mitune}, we begin with a set of fine-tuning prompts $\mathcal{Y}$ (manually crafted or already available~\citep{huang2023t2icompbench}) 
and for each prompt $y^{(i)} \in \mathcal{Y}$, 
we generate $M$ synthetic images
and record their point-wise MI (\Cref{eq:pointwise_mi} and Algorithm~\ref{algo:miestimate}). Then, we rank the image-prompt pairs $(z^{(j)}, y^{(i)}),\, j \in [1,M]$ based on the estimated point-wise MI and the top $k$ pairs to the fine-tuning dataset $\mathcal{S}$.
Finally, we fine-tune $u_\theta$ with efficient LoRA adaptation~\citep{hu2021lora}. 
Detailed fine-tuning hyperparameters in Appendix \ref{sec:finetuning_hyperparam}.


We highlight the efficiency of Algorithm~\ref{algo:miestimate} which
combines image latent generation and point-wise MI computation. Since MI estimation involves computing an expectation over diffusion times $t$, it is easy to integrate the MI estimation into the same generation loop. Moreover, the function is easy to parallelize to speed up the fine-tuning set $\mathcal{S}$ composition.

%% file: algos/mitune.tex
\IncMargin{0.8em} 




    
            

\begin{minipage}[t]{0.45\textwidth}
\fontsize{5.5}{5.5}\selectfont
\begin{algorithm}[H]

    \setcounter{AlgoLine}{0}
    \SetKwFunction{main}{PointWiseMI}
    \SetKwFunction{tk}{Top-$k$}
    \SetKwFunction{ft}{FineTune}
    \SetKwFunction{app}{append}
    \SetKwProg{Fn}{Function}{:}{}
    \SetCommentSty{mycommstyle}
    \DontPrintSemicolon
    \SetAlgoLined
    \SetKwInOut{Input}{Input}
    \SetKwInOut{Hyperparameters}{Hyperparam}
    \SetKwInOut{Output}{Output}

    \Input{Pre-trained model: $\mbu_\theta$, Prompt set: $\mathcal{Y}$} 
    \Hyperparameters{Image pool size: $M$; Top  MI-aligned samples: $k$}
    \Output{Fine-tuned diffusion model $\mbu_{\theta^*}$}
    \BlankLine

    \tcp{Fine-tuning set}
    $\mathcal{S} \leftarrow \{\,\}$   
    
    \For{$\mby^{(i)}$ in $\mathcal{Y}$}{
        \For{$j$ in $1,\cdots,M$} { 
            \tcp{Generate and compute MI}
            $\mbz^{(j)}$, $\textsc{I}(\mbz^{(j)},\mby^{(i)})$ = \main{$\mbu_\theta$, $\mby^{(i)}$}
            
            \tcp{Append samples and MI}
            $\mathcal{S}[\mby^{(i)}]$.\app{$\mbz^{(j)}, \textsc{I}(\mbz^{(j)},\mby^{(i)})$}  
        }
        \tcp{Retain only Top-$k$ elements}
        $\mathcal{S}[\mby^{(i)}]$ = \tk{$\mathcal{S}[\mby^{(i)}]$} 
    }
    \Return $\mbu_{\theta^*}$ = \ft{$\mbu_\theta$, $\mathcal{S}$}
\caption{
RFMI FT 
\label{algo:mitune}
}
\end{algorithm}
\end{minipage}
\hspace{1em}
\begin{minipage}[t]{0.5\textwidth}
\fontsize{5.5}{5.5}\selectfont
\begin{algorithm}[H]
    \SetKwFunction{main}{PointWiseMI}
    \SetKwProg{Fn}{Function}{:}{}
    \SetKwInOut{Input}{Input}
    \SetKwInOut{Output}{Output}
     \SetKwInOut{Hyperparameters}{Hyperparam}
    \DontPrintSemicolon
    \SetAlgoLined
    \SetCommentSty{mycommstyle}
    
    \Input{Pre-trained model: 
    $\mbu_\theta 
    $; 
    Prompt: 
    $\mby$}
    \Hyperparameters{Step size: $\Delta t$}     
    \Output{Generated clean latent: $\mbz$; Point-wise MI: $\textsc{I}(\mbz,\mby)$}
    \BlankLine
      
    \Fn{\main{$\mbu_\theta$, $\mby$}}{
        \tcp{Initial latent sample}
        $\mbz_0 \sim \mathcal{N}(\mathbf{0}, \mbI)$ 
        
        $\textsc{I}(\mbz,\mby)=0$
        
        \For{$t$ in $0,...,1$}{ 
            \tcp{MI estimation (\cref{eq:pointwise_mi})}
            $\textsc{I}(\mbz,\mby) 
            \mathrel{+}= 
            \frac{t}{1-t}
            u_\mbtheta (\mbz_t, \mby, t)
            \cdot
            \left( 
            u_\mbtheta (\mbz_t, \mby, t) - u_\mbtheta (\mbz_t, \emptyset, t)
            \right)
            $ 

            \tcp{Denoising step}
            $\mbz_{t} +=   u_\mbtheta (\mbz_t, \mby, t) \Delta t
                $ 
        }
    \Return $\mbz$, $\textsc{I}(\mbz,\mby)$
    }    
    \caption{
   \textsc{RFMI} 
    \label{algo:miestimate}
    }
\end{algorithm}
\end{minipage}

%% file: sections/experiments.tex
\vspace{-0.2cm}
\section{Experimental Evaluation}

\subsection{Synthetic benchmark}\label{sec:exp_1}

As a preliminary step, we assess the quality of RFMI
using a known 
benchmark \citep{czyż2023normalevaluationmutualinformation}
composed of 40 tasks with
synthetic data generated from a variety of known
distributions where the true MI is known, and venturing beyond 
the typical Gaussian distributions
to include harder cases (e.g.,
distributions with high MI or long tails).

We consider four alternative neural estimators as baselines,
namely MINE \citep{mine}, 
InfoNCE \citep{infonce},
NWJ \citep{nwj} and 
DOE \citep{mcallester2020formallimitationsmeasurementmutual}.
All methods are trained/tested using 100k/10k samples,
where each sample is composed of two data points 
$x$ and $y$ concatenated as input for the neural network.

\begin{figure}[!t]
\centering
\includegraphics[width=\textwidth]{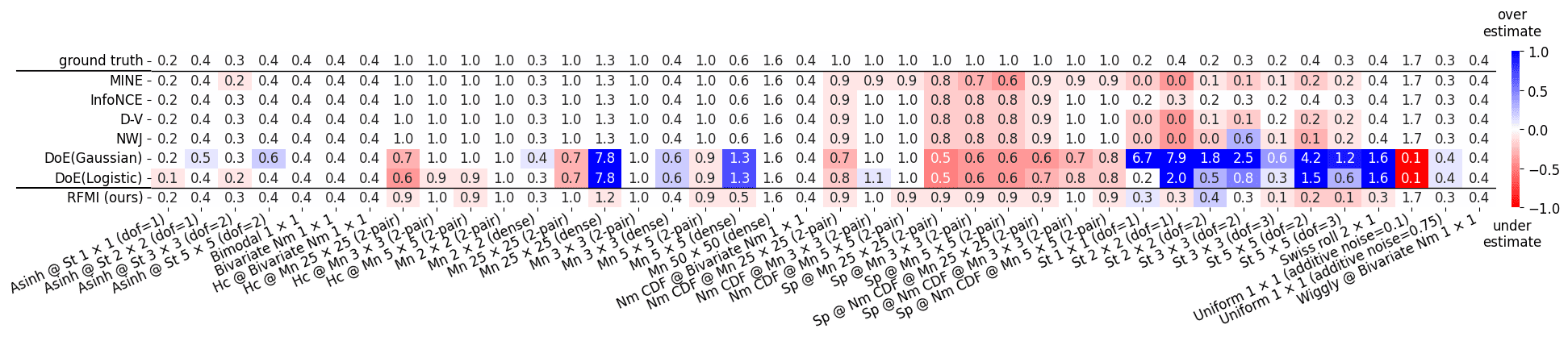}
\caption{MI estimation results. Color indicates relative negative bias (red) and positive bias (blue). 
}
\label{fig:mi_heatmap}
\end{figure}

Figure~\ref{fig:mi_heatmap} shows the ground truth MI
and each method estimates, with colors reflecting the difference
between the MI estimate and the true value -- the lighter the shade, the smaller
the estimation error. Overall, RFMI is on par or better than alternative methods.

\begin{table}[t!]
\small
\centering
\captionof{table}{T2I-CompBench alignment results ($\%$) on images generated with CFG=4.5
}
\label{tab:T2I-CompBench}
\begin{tabular}{
    @{$\:\:\:$}
    r
    @{$\:\:\:$}
    c
    @{$\:\:\:$}
    c
    @{$\:\:\:$}
    c
    @{$\:\:\:$}
    c
}
\toprule
Category & Shape & 2D-spatial & 3D-spatial & Numeracy
\\
        & (BLIP-VQA) 
        & (UniDet) 
        &  (UniDet)\footnotemark[2]
        & (UniDet)
\\
\midrule
SD3.5-M 
&  
57.96 
& 31.31 & 38.81 & 61.15
\\
RFMI FT
& 
61.78 
& 33.92 & 42.28 & 64.00
\\
\midrule
\textit{abs. difference}
& 3.82 & 2.61 & 3.47 & 2.85
\\
\textit{relative gain}
& 6.59 & 8.34 & 8.94 & 4.66
\\
\bottomrule
\end{tabular}
\vspace{-0.3cm}
\end{table}
\footnotetext[2]{
        For prompts depicting 3D-spatial relation, T2I-CompBench++ leverages UniDet~\citep{unidet} for object detection and Dense Vision Transformer~\citep{ranftl2021visiontransformersdenseprediction}
        for depth estimation.
        } 

\subsection{Text-Image Alignment Evaluation}\label{sec:exp_2}

We evaluate RFMI on T2I-CompBench++~\citep{huang2023t2icompbench}, 
a T2I benchmark  composed of 700/300 (train/test) prompts across 8 categories including 
attribute binding (color, shape, and texture categories), 
object relationships (2D-spatial, 3D-spatial, and non-spatial associations), 
numeracy
and complex composition tasks. 
Prompts are generated with predefined rules or ChatGPT~\citep{chatgpt} and
the evaluation uses
BLIP-VQA~\citep{huang2023t2icompbench}, UniDet~\citep{unidet}, or GPT-4V~\citep{gpt4v}
depending on the category.
As~\citet{huang2023t2icompbench} recently found, Stable Diffusion 3 already ``saturates'' 
performance on certain categories (see Table XIII in \cite{huang2023t2icompbench}): then, 
we focus our evaluation only on categories having an alignment score lower than 0.7, namely
the 4 categories shape, 2D-spatial, 3D-spatial, and numeracy.


We applied RFMI FT (Algorithm \ref{algo:mitune}) on Stable Diffusion 3.5-Medium~\cite{SD3} (SD3.5-M)\footnote[3]{From a preliminary investigation we observed a $\times4$ computational costs for FLUX~\citep{flux2023} so we could not collect results in time for the submission.} and extended Huang et al. benchmark to include this latest RF-based T2I model. 
Table~\ref{tab:T2I-CompBench} collects the results, with absolute difference and percentage gain 
between SD3.5-M and RFMI FT 
summarized at the bottom.
As expected, 
RFMI FT improves the T2I alignment of SD3.5-M by a sizable margin across all the $4$ challenging categories. Qualitative visualization examples are shown in \Cref{fig:visu}.

We highlight that our approach RFMI FT is not sensitive to the neural network architecture or the type of data, so it could be integrated beyond T2I task and into other disciplines where rectified flow is adopted for conditional generation.

\begin{minipage}[t]{1.0\textwidth}
\fontsize{7}{7}\selectfont
\begin{tabular}{
@{}
p{63pt}@{}
p{63pt}@{}
p{8pt}@{}
p{63pt}@{}
p{63pt}@{}
p{8pt}@{}
p{63pt}@{}
p{63pt}@{}
}
\hfil\fontsize{5.8}
{5.8}\selectfont{SD3.5-M} &
\hfil\fontsize{5.8}
{5.8}\selectfont{RFMI FT} &
&
\hfil\fontsize{5.8}
{5.8}\selectfont{SD3.5-M} &
\hfil\fontsize{5.8}
{5.8}\selectfont{RFMI FT} &
&
\hfil\fontsize{5.8}
{5.8}\selectfont{SD3.5-M} &
\hfil\fontsize{5.8}
{5.8}\selectfont{RFMI FT} 
\\
\end{tabular}
\includegraphics[width=0.32\linewidth]
{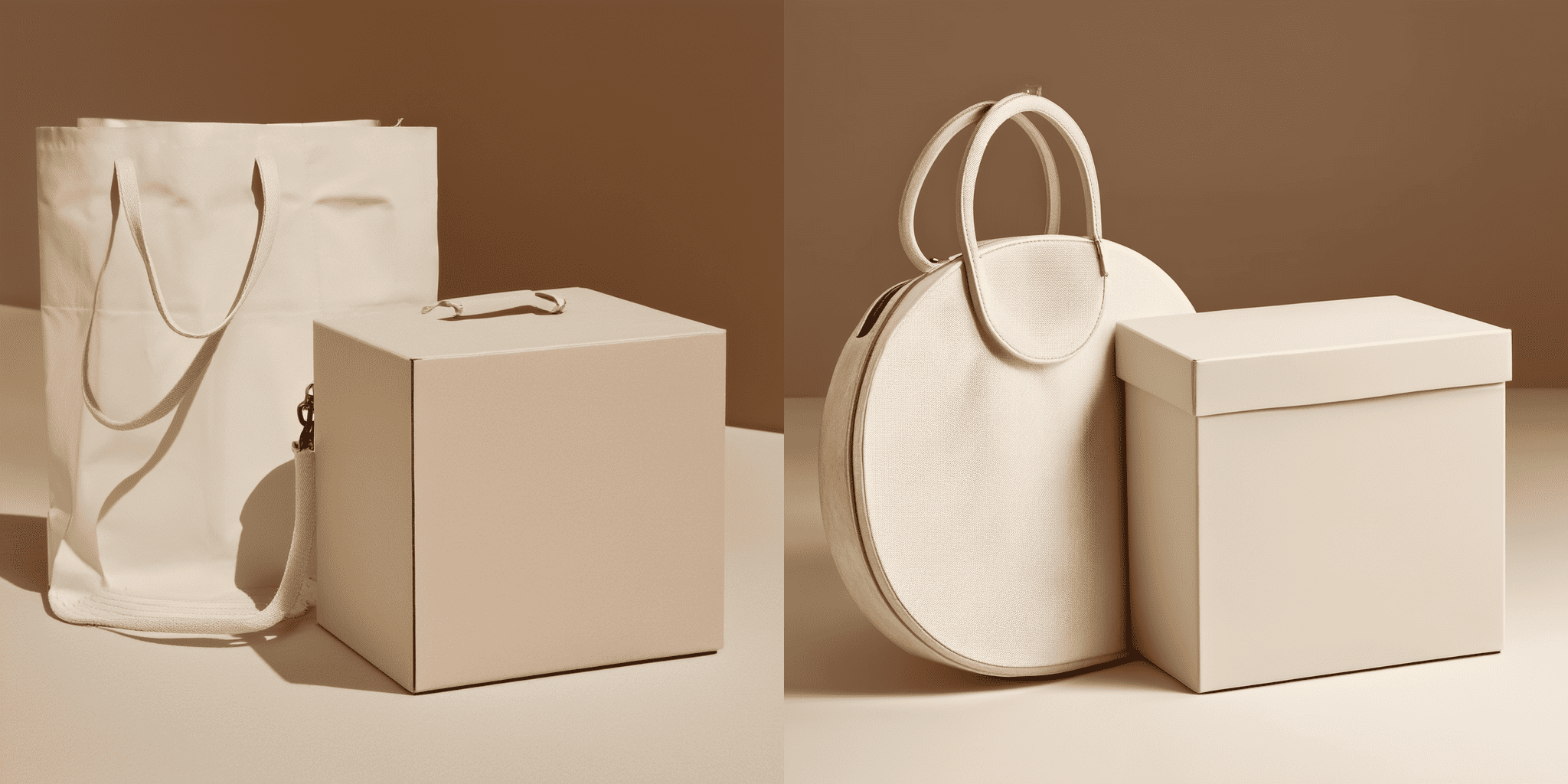}
\hspace{2pt}
\includegraphics[width=0.32\linewidth]{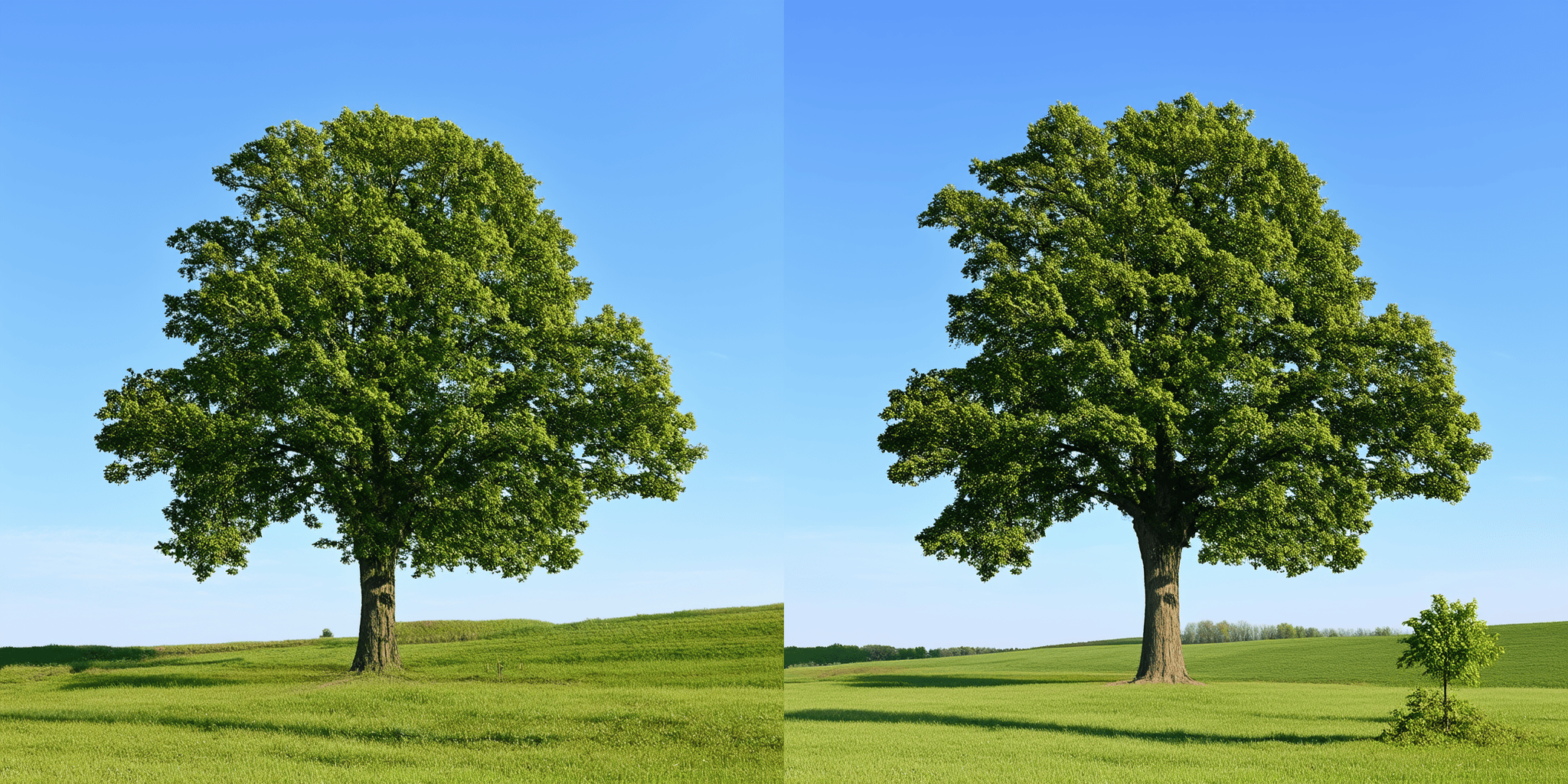}
\hspace{2pt}
\includegraphics[width=0.32\linewidth]{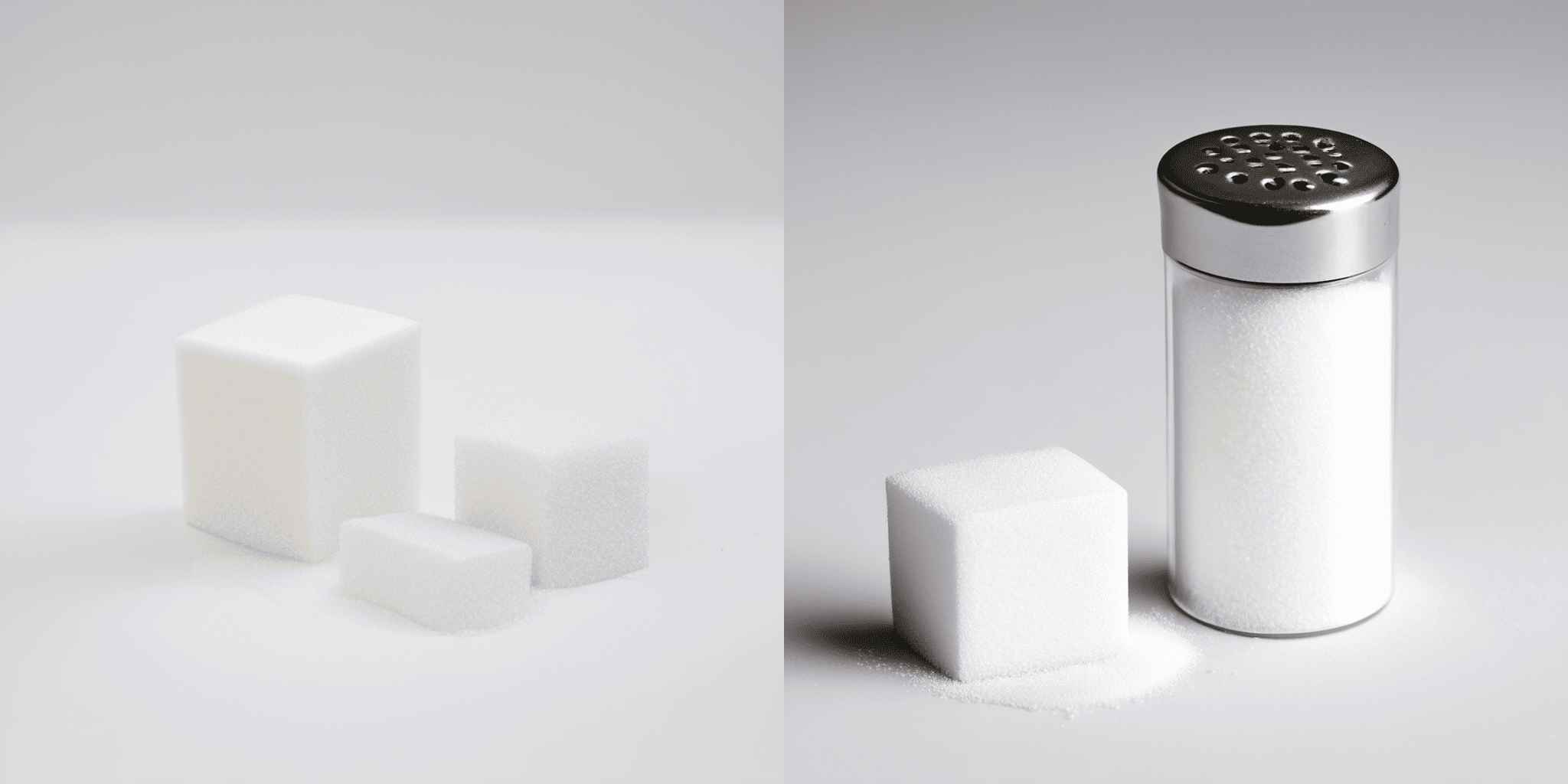}
\\
\fontsize{5.5}{5.5}\selectfont
(Shape)
``a round bag and a square box''
\hspace{18pt} 
``a tall oak tree and a short sapling''
\hspace{3pt} 
``a cubic sugar cube and a cylindrical salt shaker'' 

\vspace{3pt}

\includegraphics[width=0.32\linewidth]
{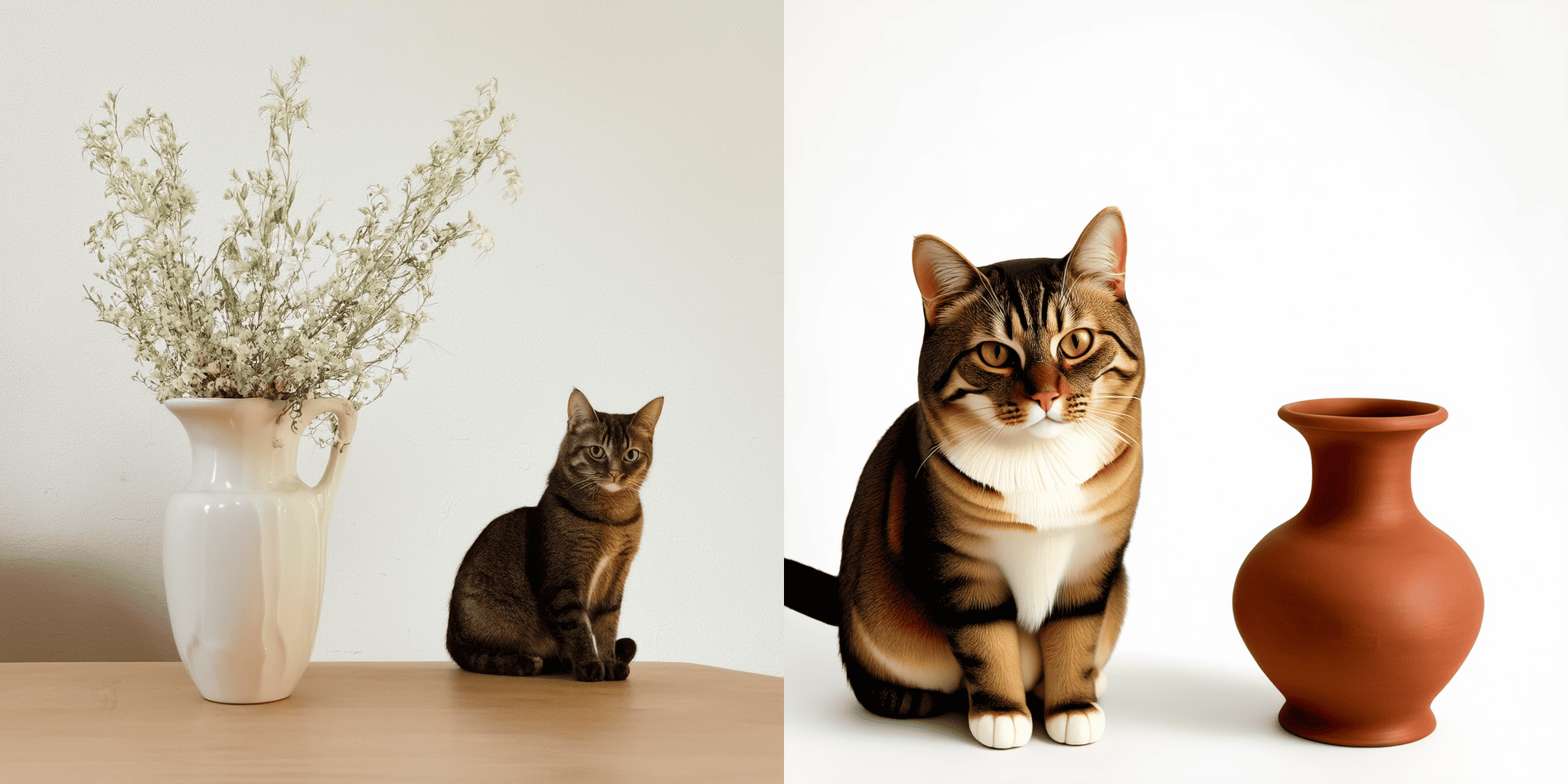}
\hspace{2pt}
\includegraphics[width=0.32\linewidth]
{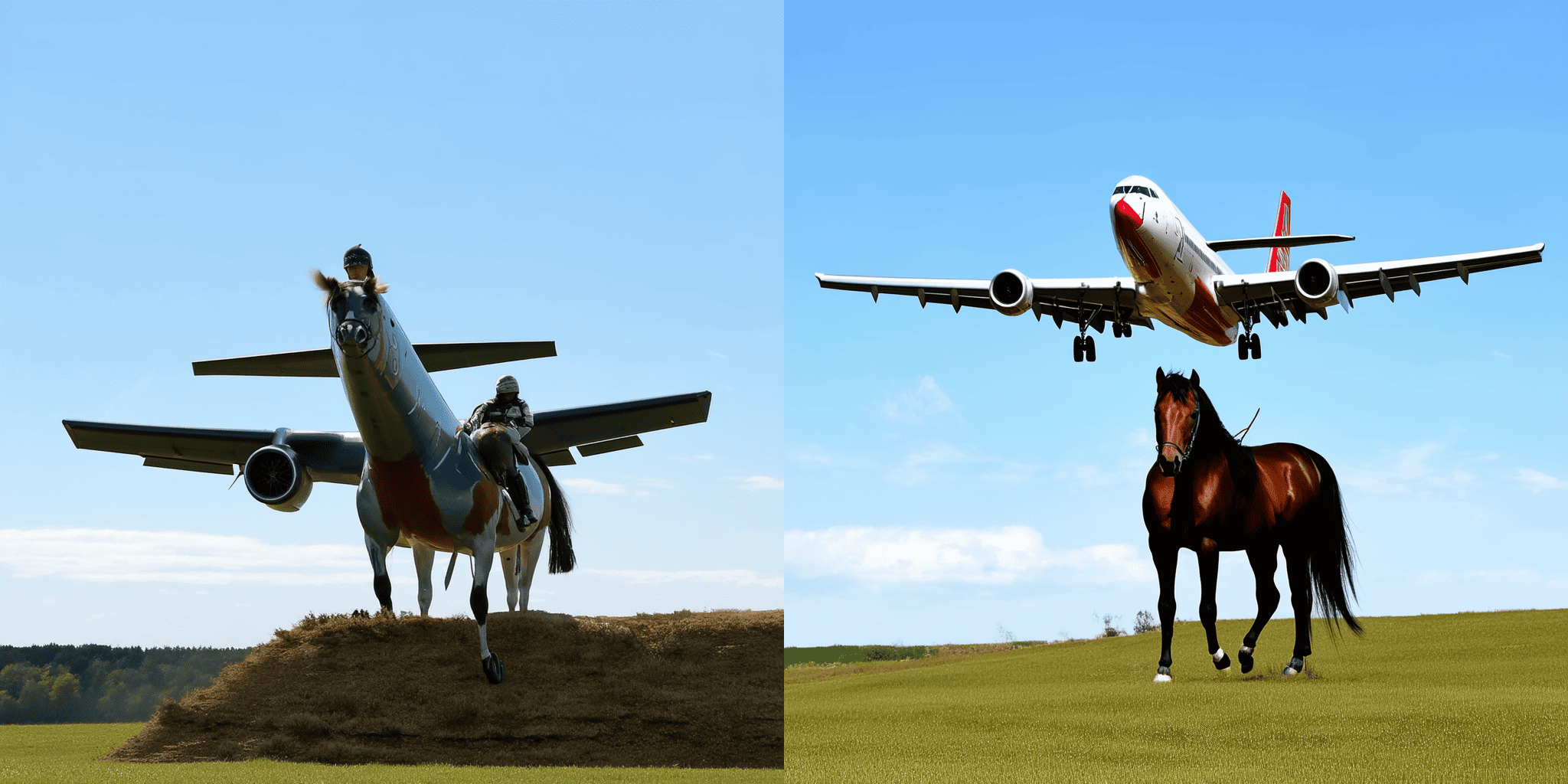}
\hspace{2pt}
\includegraphics[width=0.32\linewidth]{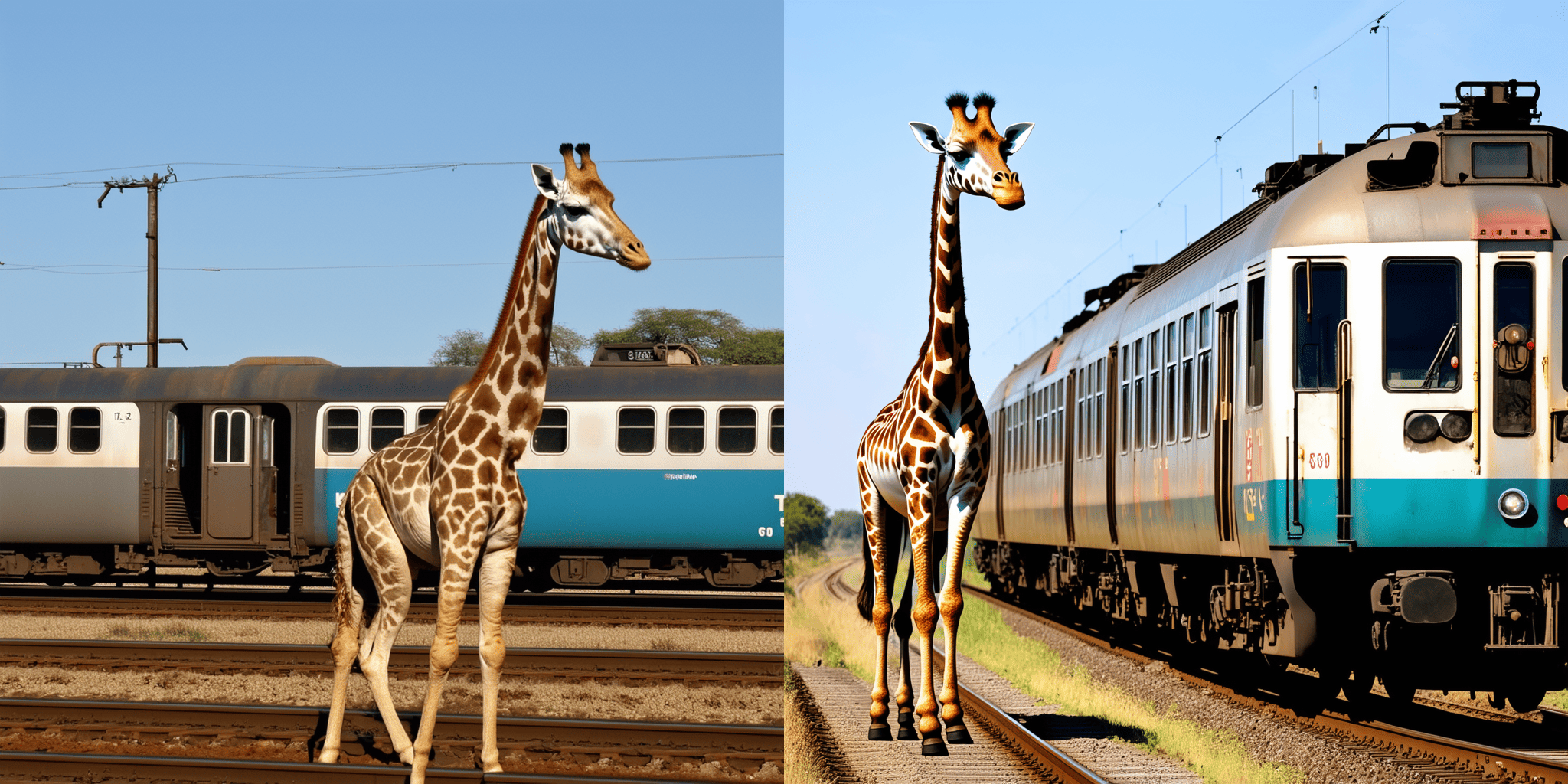}
\\
\fontsize{5.5}{5.5}\selectfont
(2D-spatial)
``a vase on the right of a cat''
\hspace{18pt} 
``a airplane on the top of a horse''
\hspace{30pt} 
``a giraffe on the left of a train'' 

\vspace{3pt}

\includegraphics[width=0.32\linewidth]
{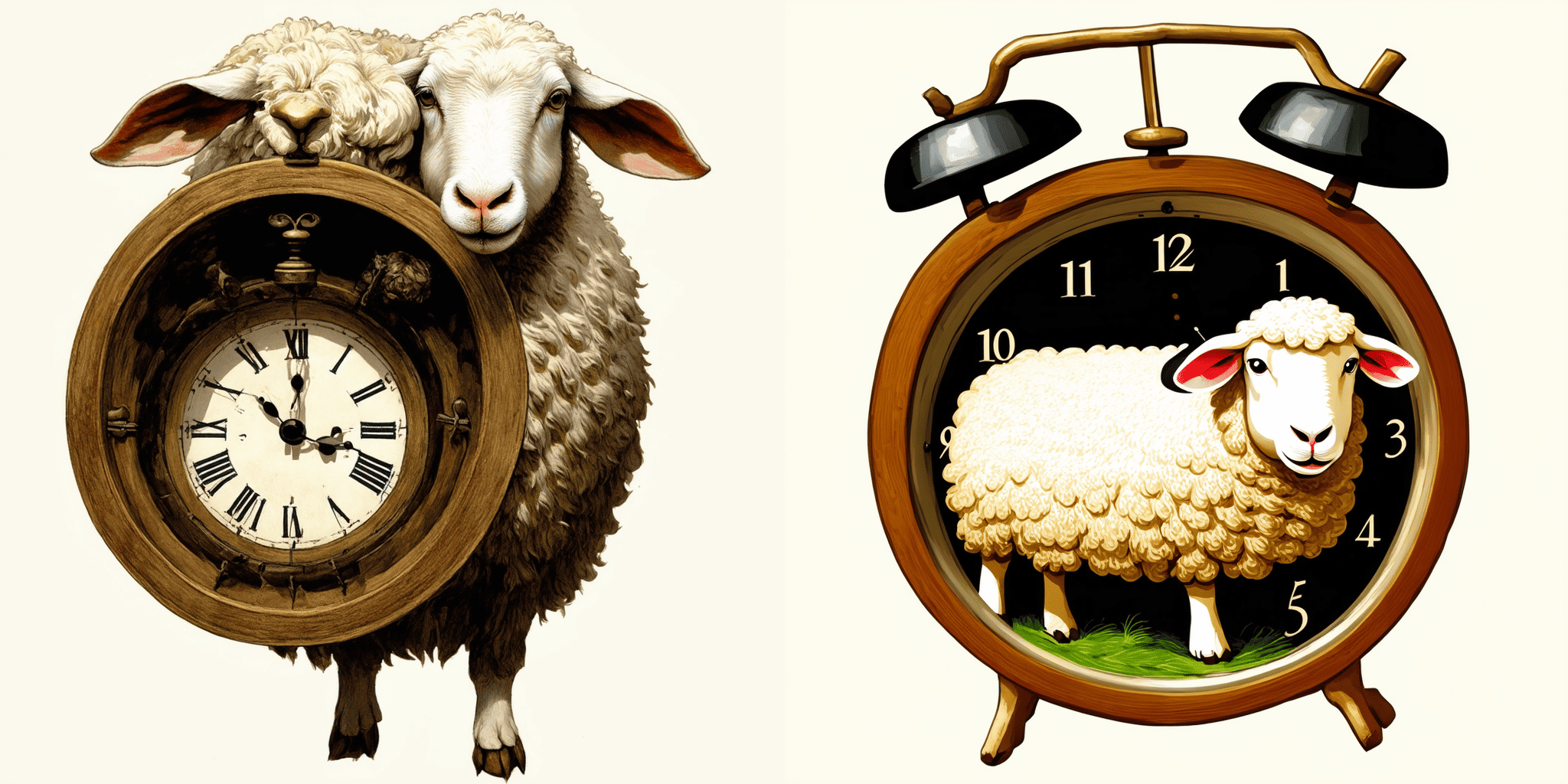}
\hspace{2pt}
\includegraphics[width=0.32\linewidth]
{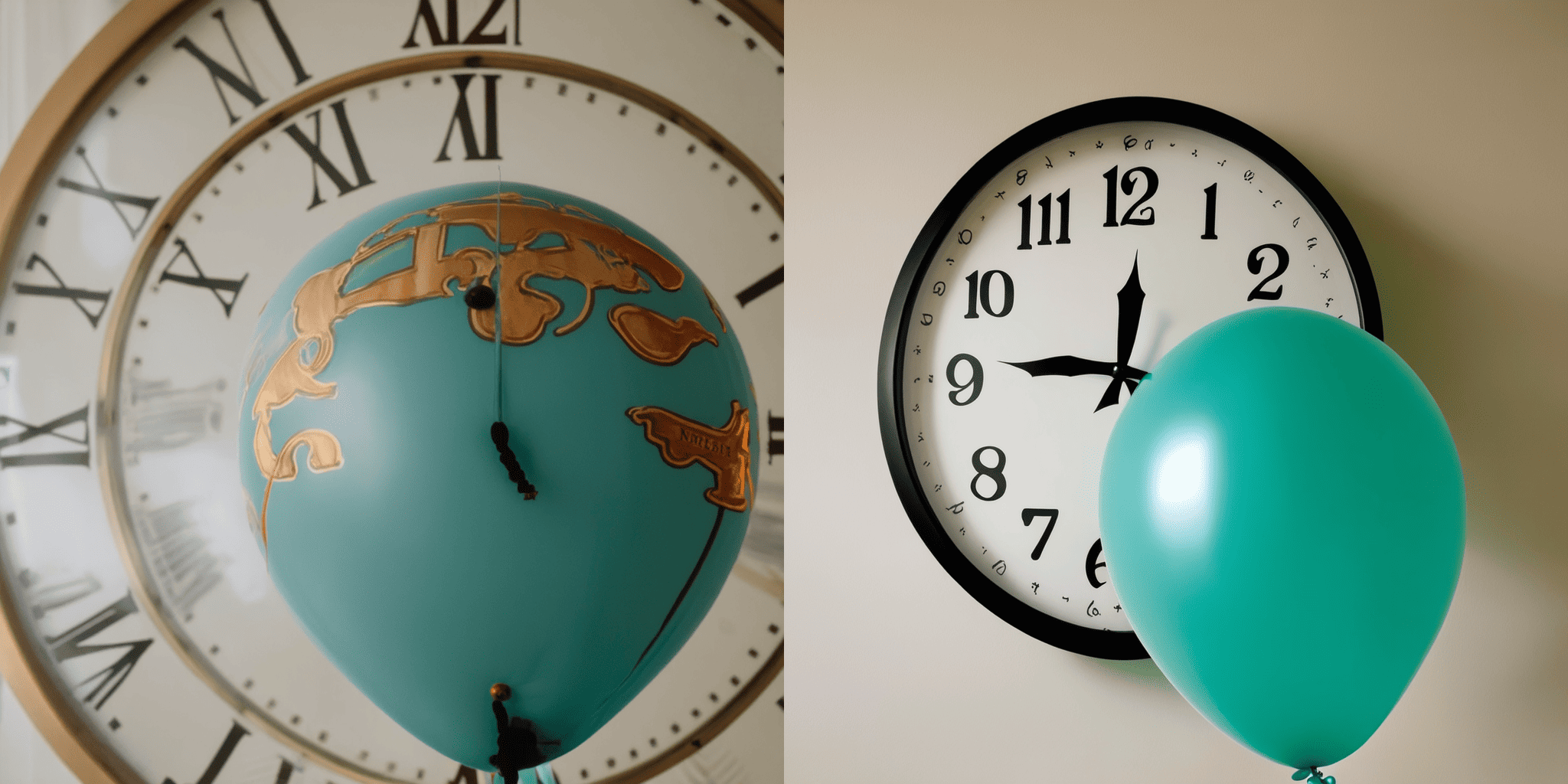}
\hspace{2pt}
\includegraphics[width=0.32\linewidth]{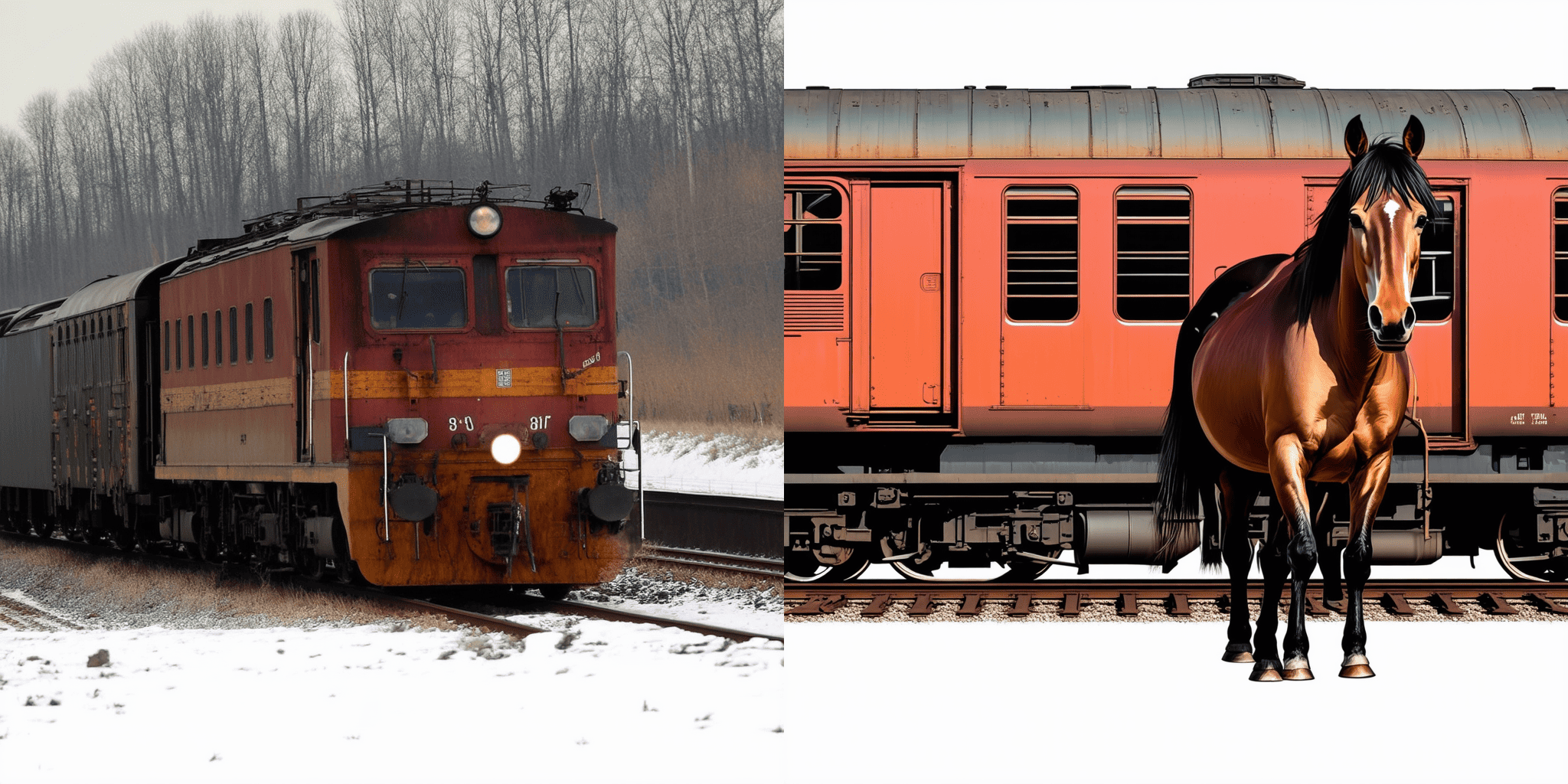}
\\
\fontsize{5.5}{5.5}\selectfont
(3D-spatial)
``a clock hidden by a sheep''
\hspace{20pt} 
``a balloon in front of a clock''
\hspace{35pt} 
``a train hidden by a horse'' 

\vspace{3pt}

\includegraphics[width=0.32\linewidth]
{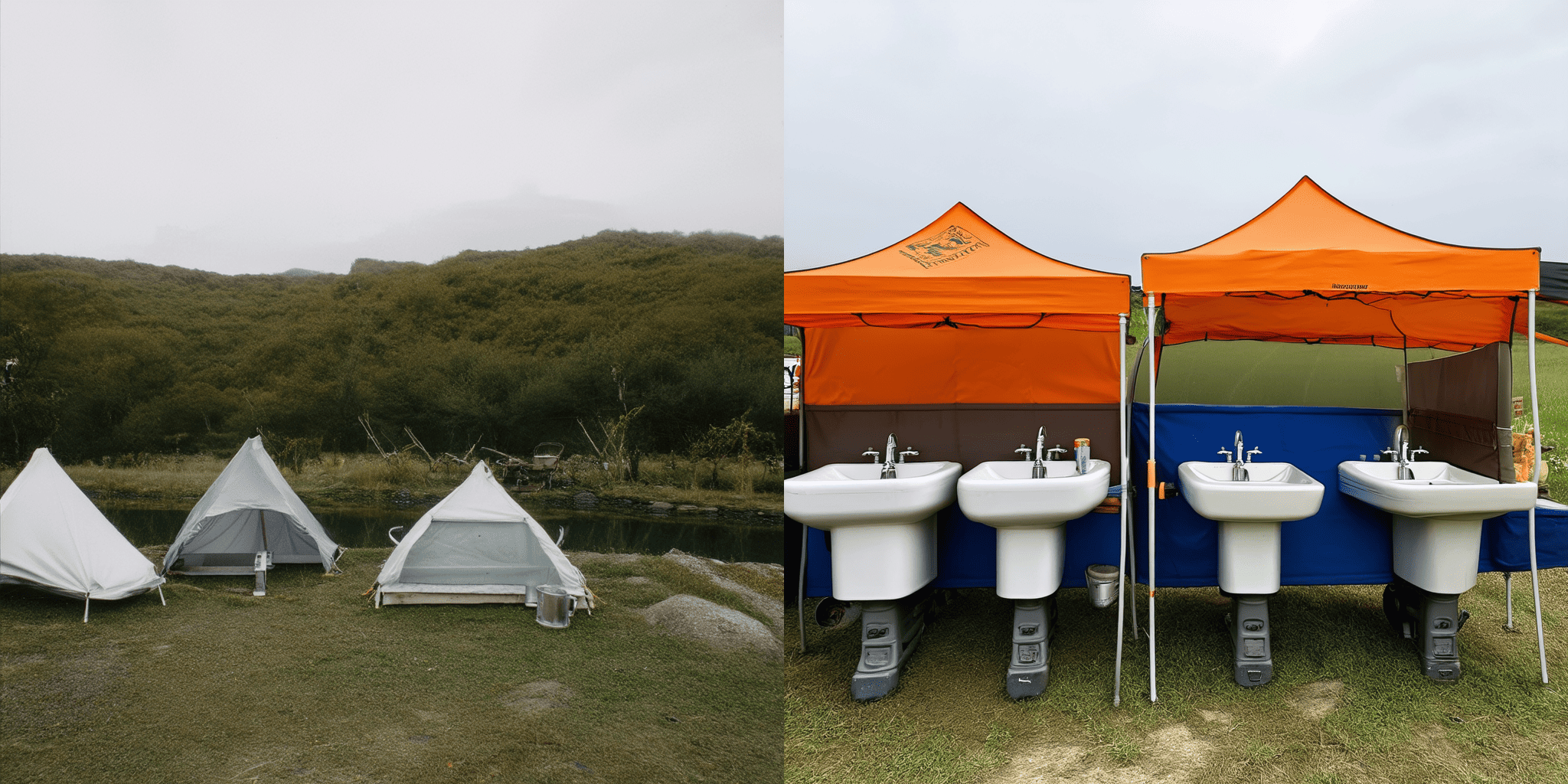}
\hspace{2pt}
\includegraphics[width=0.32\linewidth]
{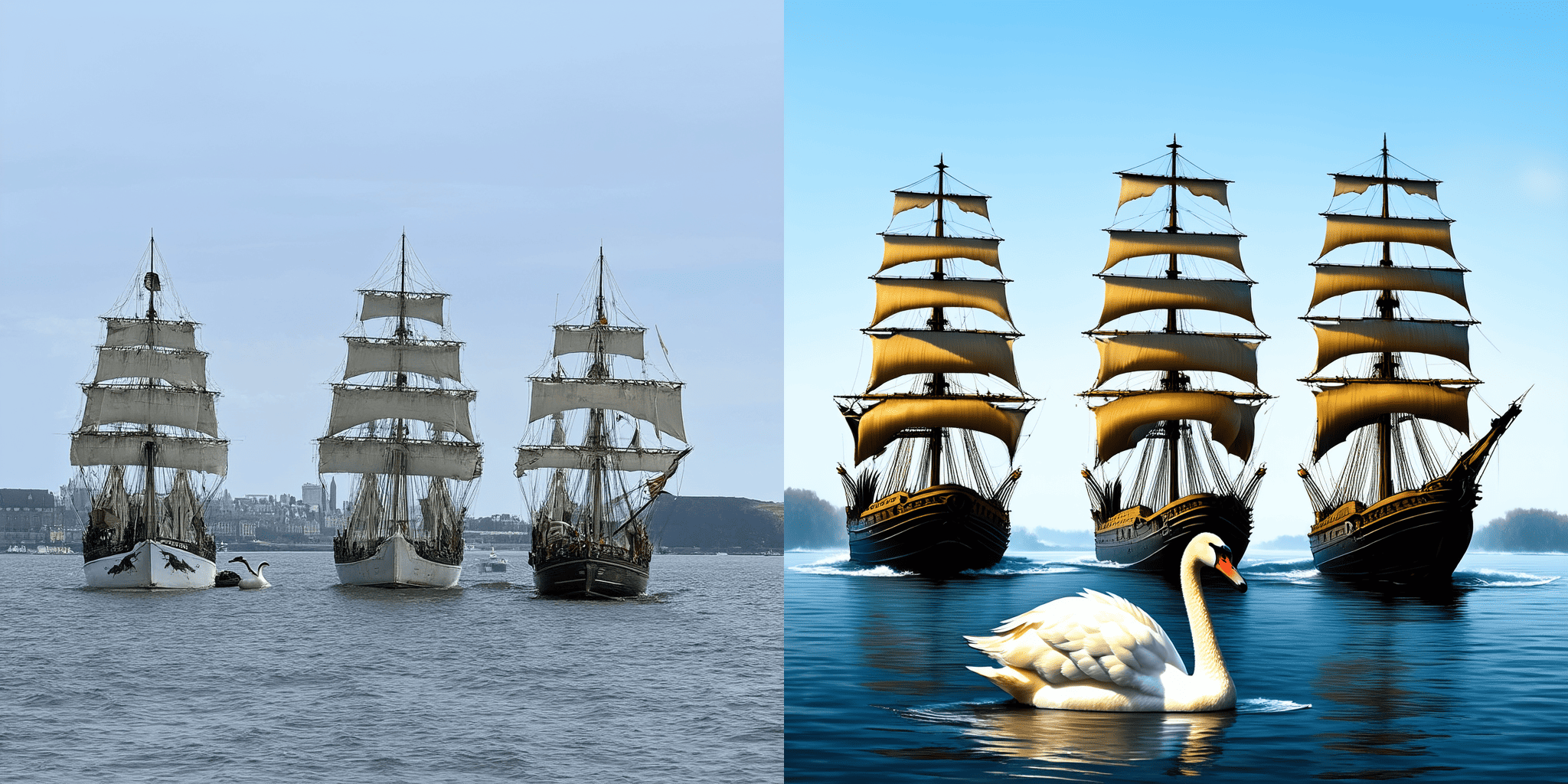}
\hspace{2pt}
\includegraphics[width=0.32\linewidth]{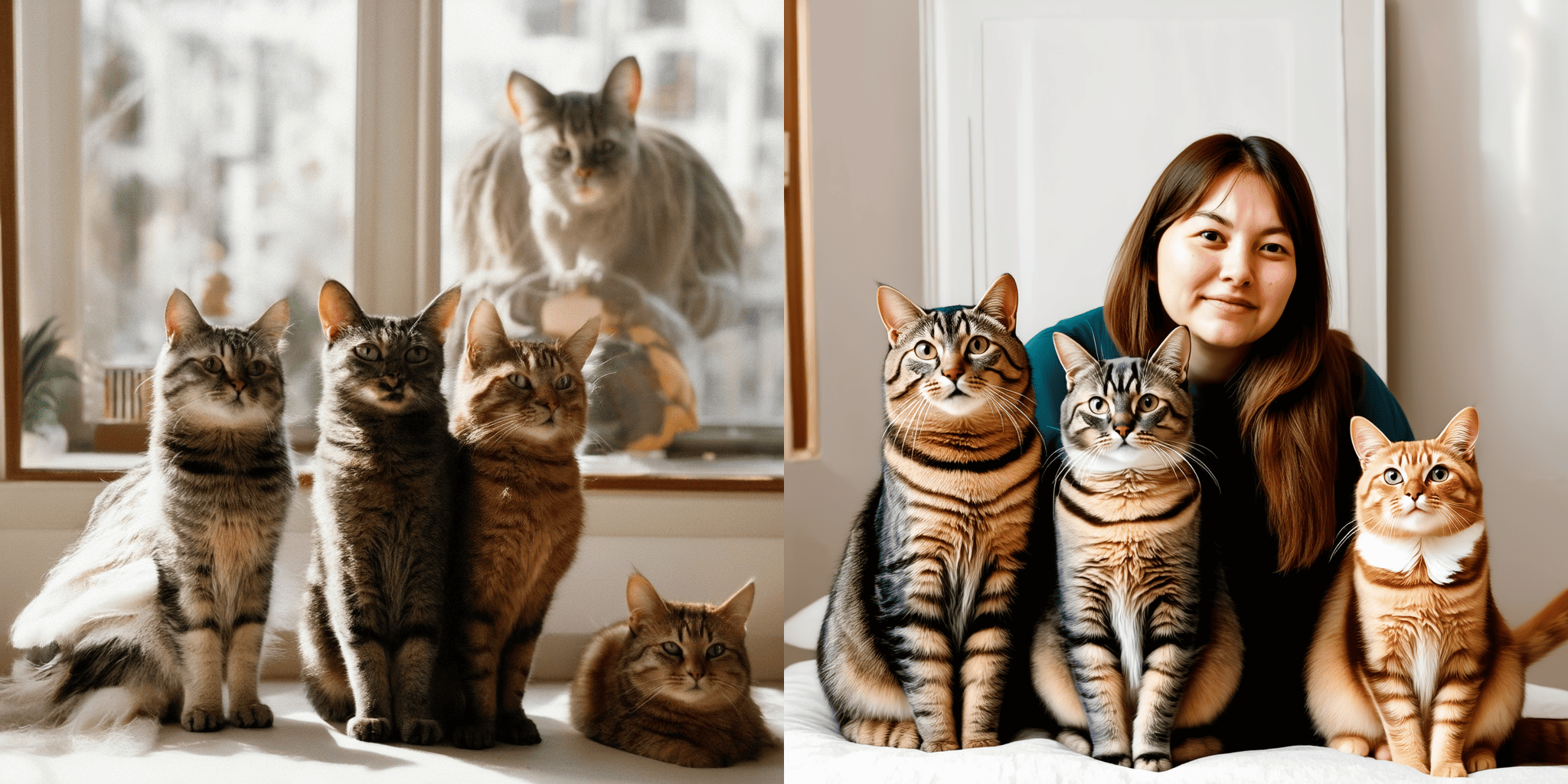}
\\
\fontsize{5.5}{5.5}\selectfont
(Numeracy)
``four sinks and two tents''
\hspace{20pt} 
``three ships sailed alongside one swan''
\hspace{45pt} 
``one person and three cats''

\vspace{-3pt}
\centering
\captionof{figure}{
Qualitative examples from \Cref{tab:T2I-CompBench} (same seed used for a given prompt).
}
\label{fig:visu}

\end{minipage}




%% file: sections/conclusion.tex
\vspace{-0.2cm}
\section{Conclusion}

\vspace{-0.2cm}
In this study, we introduced RFMI, a novel RF-based MI estimator which provides a unique perspective on MI estimation by leveraging the theory of FM-based generative models.
To show its effectiveness, we first considered a synthetic benchmark
where the true MI is known and we showed that RFMI is on par or better than alternative neural estimators.
Then, we considered the T2I alignment problem and used RFMI in a
self-supervised fine-tuning approach. Specifically, 
we used the point-wise MI value between text and image estimated by the pre-trained RF model to create a synthetic fine-tuning set for improving the model alignment.
Our empirical evaluation on MI estimation benchmark and T2I alignment benchmark illustrated the effectiveness of RFMI.
Our lightweight, self-supervised fine-tuning method does not depend on specific model architectures, so it can be used to improve alignment of a variety of RF models in the future.

%% file: sections/proofs_v1.tex
\subsection{Proofs}


\subsubsection{Proof of the limiting case of proposition~\ref{prop:rho-v}}\label{sec:proof_lemma1}

\begin{proposition}
  (Limiting case of Prop.~\ref{prop:rho-v}) 
    For linear conditional flow with Gaussian prior, 
    as $t\rightarrow1$, the relation between the marginal velocity field and the score function of the marginal probability path is:
\begin{equation}
  \begin{cases}
      \begin{aligned}
   \lim_{t \rightarrow 1}  \nabla \log p_{t \mid Y}(x \mid y) &
  = 
  \lim_{t \rightarrow 1} 
   -
    \partial_t u_t(x \mid y )
    \\
    \lim_{t \rightarrow 1}   \nabla \log p_{t}(x) &=  
     \lim_{t \rightarrow 1} 
     -
    \partial_t u_t(x)
       \end{aligned}
       \end{cases}
\end{equation}
\end{proposition}




\textit{Proof.} 
To simplify notation, we present proof in the \textit{unconditional} setting.
Consider the case of conditional flow at the form $\psi_t\left(x \mid x_1\right)=a_t x + b_t x_1$, 
where $a_t$ and $b_t$ are chosen to
satisfy 
$\psi_\iota\left(x \mid x_1\right)= \begin{cases}x & t=0 \\ x_1 & t=1\end{cases}$,  
$x$ is sample $x_0$ of RV $X_0 \sim p$,
and 
$x_1$ is sampled from the target distribution $q$.
The marginalization trick shows that 
$\psi_t$ generates a $p_t$ satisfying $p_0=p$ and $p_1 = q$
. 
Using the expression of marginal probability flux (Eq. (14) in ~\cite{albergo2023buildingnormalizingflowsstochastic}), 
at $t = 1$, we have:
\begin{equation}
\begin{aligned}
j_{t=1}(x)
&= 
\int_{\mathbb{R}^d \times \mathbb{R}^d} 
\left[
\partial_t 
\psi_t
\left(x_0 | x_1\right) 
\right]
|_{t=1}
\delta\left(x-
\psi_{t=1}
\left(x_0 | x_1\right) 
\right) p_0\left(x_0\right) p_1\left(x_1\right) d x_0 d x_1 \\
&= 
\int_{\mathbb{R}^d \times \mathbb{R}^d} 
\left( \dot{a}_t |_{t=1} x_0+\dot{b}_t |_{t=1} x_1     \right)
\delta\left(x-x_1\right) 
p_0\left(x_0\right) p_1\left(x_1\right) d x_0 d x_1 \\  
&= 
\int_{x_0 \in \mathbb{R}^d}   x_0  p_0\left(x_0\right) d x_0 
\int_{x_1 \in \mathbb{R}^d} 
 \dot{a}_t |_{t=1} 
p_1\left(x_1\right) \delta\left(x-x_1\right) 
d x_1 
 \quad + \\
& \quad 
\int_{x_0 \in \mathbb{R}^d} p_0\left(x_0\right) d x_0  
\int_{x_0 \in \mathbb{R}^d}  \dot{b}_t |_{t=1} x_1 p_1\left(x_1\right) \delta\left(x-x_1\right) d x_1 \\
&=  
\mathbb{E}[X_0]  \dot{a}_1 
p_1\left(x_1\right) 
+ 
\dot{b}_1 x_1 
p_1\left(x_1\right) 
\\
\end{aligned}
\end{equation}

It follows that the marginal velocity field at $t=1$ is 
\begin{equation}
\label{eq:v1}
\begin{aligned}
   u_1(x) & = j_1(x) / p_1(x) \\
   & = 
   \mathrm{E}[X_0]  \dot{a}_1 
+ 
\dot{b}_1 x_1 
\end{aligned}
\end{equation}

If the conditional flow is linear, we have $a_t = (1-t)$ and  $b_t = t$, and therefore $\dot{a}_t = -1$, $\dot{b}_t = 1$.
Furthermore, if $X_0$ is the standard Gaussian, we have $ \mathrm{E}[X_0] = 0$.
This means that
\Cref{eq:v1} becomes 
$
u_1(x) =   0   \times (-1)
+ 
1 \times x_1 = x_1
$. Inserting this
equality into the numerator in \Cref{eq:score2}, 
both the denominator and the numerator converge to $0$ when  $t \rightarrow 1$:
\begin{equation}
    \begin{cases}
        t u_t(x) - x
\xrightarrow[]{t\rightarrow1} 1 \times x_1 - x_1 = 0 \\
1-t 
\xrightarrow[]{t\rightarrow1} 1-1=0
    \end{cases}
\end{equation}
By applying l'Hôpital's rule, 
$$
\begin{aligned}
& \nabla  log p_1(x) \\
& =  
\lim _{t \rightarrow 1}
 \frac
 { 
\partial_t 
 \left(t u_t(x) - x \right)
 = [\partial_t t] u_t(x) + t [\partial_t u_t(x) ]
  - [\partial_t x ]
 =  u_t(x) +  t  [\partial_t u_t(x) ] - u_t(x)
 }
 { \partial_t  (1-t) = -1} \\
 & = \lim _{t \rightarrow 1} [-\partial_t u_t(x)] 
\end{aligned} 
$$

It is easy to show that the considerations above also hold in the \textit{guided} case, whereby 
the marginal items 
(i.e., the probability flux $j_{t}$, flow $\psi_{t}$, velocity field $u_t$, 
probability path $p_{t}$ and target distribution $q$)
are expressed in their guided form. 
$\quad \quad \square$


\subsubsection{Proof of \Cref{prop:micomp}}
\label{sec:proof_prop2}


\begin{proposition}[MI computation]
{
Given a linear conditional flow with Gaussian prior, the MI between the target data $X$ and the guidance signal $Y$ is given by 
\begin{equation}\label{eq:mi_proof}
\begin{aligned}
\mathrm{I}(X ; Y)
&=
\mathbb{E}_Y\left[
    \int_{\mathbb{R}^d} 
    p_{1 \mid Y}(x_1 \mid Y)
    \log \left(\frac{
    p_{1 \mid Y}(x_1 \mid Y)
    }{
    p_{1}(x_1) 
    }\right) \mathrm{d} x_1
\right] 
\\
&=
\mathbb{E}_Y\left[
   \int_0^1 
        \mathbb{E}_{X_t \mid Y}
        \left[   
            \frac{t}{1-t}
            u_t(X_t|Y)  \cdot
                \left( u_t(X_t|Y) - u_t(X_t) 
                \right)
        \right]
    dt
\right]. 
\end{aligned}
\end{equation}

}
\end{proposition}


\textit{Proof.} 
What needs to be proved in~\Cref{eq:mi_proof} is the equivalence of the terms inside the expectation. 
To keep notation concise, in the following we will rename 
the guided pair  
$\left(  p_{t \mid Y}(\cdot \mid y), u_t(\cdot \mid y)  \right)$
as simply 
$\left( p_t^{A}(\cdot), u_t^{A}(\cdot) \right)$, 
and the marginal pair  
$\left(  p_{t}(\cdot), u_t(\cdot)  \right)$
as  
$\left( p_t^{B}(\cdot), u_t^{B}(\cdot) \right)$, 
so what needs to be proved becomes:
\begin{equation}
\label{eq:to_prove}
  \int_{\mathbb{R}^d}  
    p_{1}^A(x_1)
    \log \left(\frac{
    p_{1}^A(x_1)
    }{
   p_{1}^B(x_1)
    }\right) \mathrm{d} x_1
=
\int_0^1 
\mathbb{E}_{p_t^A} 
 \left[   
   \frac{t}{1-t}
u_t^A(x)  \cdot
  \left( u_t^A(x) - u_t^B(x) \right)
  \right]
dt
\end{equation}
To prove \Cref{eq:to_prove}, we start with 
expanding its LHS:
\begin{equation}\label{eq:term1and2}
\begin{aligned}
  &\int_{\mathbb{R}^d}  
    p_{1}^A(x)
    \log \left(\frac{
    p_{1}^A(x)
    }{
   p_{1}^B(x)
    }\right) \mathrm{d} x \\
& \stackrel{(i)}{=}
  \int_{\mathbb{R}^d}  
    p_{0}^A(x)
    \log \left(\frac{
    p_{0}^A(x)
    }{
   p_{0}^B(x)
    }\right) \mathrm{d} x
+
\int_0^1
\partial_t 
  \int_{\mathbb{R}^d}  
    p_{t}^A(x)
    \log \left(\frac{
    p_{t}^A(x)
    }{
   p_{t}^B(x)
    }\right) \mathrm{d} x
dt \\
& \stackrel{(ii)}{=} 0+
\int_0^1
\partial_t 
  \int_{\mathbb{R}^d}  
    p_{t}^A(x)
    \log \left(\frac{
    p_{t}^A(x)
    }{
   p_{t}^B(x)
    }\right) \mathrm{d} x
dt \\
& \stackrel{(iii)}{=}
\int_0^1 \int_{\mathbb{R}^d}
\partial_t
 \left[ p_t^A\left(x\right) \log \left(\frac{p_t^A\left(x\right)}{p_t^B\left(x\right)}\right)\right] \mathrm{d} x
dt \\
& \stackrel{(iv)}{=}
\int_0^1 
\left[
\underbrace{
 \int_{\mathbb{R}^d}  \left[\partial_t
 p_t^A\left(x\right) \right] \left[\log \left(\frac{p_t^A\left(x\right)}{p_t^B\left(x\right)}\right) \right] \mathrm{d} x
 }_{\text{\textcircled{1}}}
 + 
 \underbrace{
 \int_{\mathbb{R}^d} \left[
 p_t^A\left(x\right) \right] \left[\partial_t\log \left(\frac{p_t^A\left(x\right)}{p_t^B\left(x\right)}\right) \right] \mathrm{d} x
  }_{\text{\textcircled{2}}}
 \right] dt
\end{aligned}
\end{equation}

where $(i)$ follows from the fundamental theorem of calculus;
$(ii)$  follows from 
the fact that
both 
$p_0^A\left(x\right)$ 
and 
$p_0^B\left(x\right)$ 
coincide with source distribution $p$ at $t=0$;
$(iii)$ follows from 
switching differentiation ($\partial_t$) and integration ($\int_{\mathbb{R}^d}$) as justified by
Leibniz’s rule;
$(iv)$ follows from using the Product Rule (i.e. $(u \cdot v)^{\prime}=u^{\prime} \cdot v+u \cdot v^{\prime}$) on $\partial_t$.



About the term \textcircled{1} in \Cref{eq:term1and2},
using 
in sequential order
$(i)$ the Continuity Equation 
on $[\partial_t
 p_t^A\left(x\right)]$, 
 $(ii)$ the Product Rule, 
 $(iii)$ the Divergence Theorem,
and 
$(iv)$ the assumption that $ p_t^A$ vanishes at infinity 
 gives
\begin{equation}\label{eq:term1}
\begin{aligned}
& 
\int_{\mathbb{R}^d} \left[\partial_t
 p_t^A\left(x\right) \right]\left[\log \left(\frac{p_t^A\left(x\right)}{p_t^B\left(x\right)}\right)  \right] \mathrm{d} x \\
\stackrel{(i)}{=}&
 \int_{\mathbb{R}^d}\left[-\nabla_x \cdot ( p_t^A(x) u_t^A(x)  ) \right]\left[\log \left(\frac{p_t^A\left(x\right)}{p_t^B\left(x\right)}\right)  \right]\mathrm{d} x \\
\stackrel{(ii)}{=}&
 \int_{\mathbb{R}^d} \left( p_t^A(x) u_t^A(x)  \right) \cdot \left[\nabla_x  \log \left(\frac{p_t^A\left(x\right)}{p_t^B\left(x\right)}\right) \right]\mathrm{d} x   
-
 \int_{\mathbb{R}^d}
\nabla_x \cdot 
\left( 
p_t^A(x) u_t^A(x)
\log \left(
\frac{p_t^A\left(x\right)}{p_t^B\left(x\right)} 
 \right)
 \right)
 dx
\\
\stackrel{(iii)}{=}&
 \int_{\mathbb{R}^d}  \left( p_t^A(x) u_t^A(x)  \right) \cdot \left[\nabla_x  \log \left(\frac{p_t^A\left(x\right)}{p_t^B\left(x\right)}\right) \right]\mathrm{d} x   
-
 \oint_{\partial \mathbb{R}^d}
\left( 
p_t^A(x) u_t^A(x)
\log \left(
\frac{p_t^A\left(x\right)}{p_t^B\left(x\right)} 
 \right)
 \right)
 \cdot \mathbf{n} d S
\\
 \stackrel{(iv)}{=}&
 \int_{\mathbb{R}^d} \left( p_t^A(x) u_t^A(x)  \right) \cdot
   \left( \nabla_x  \log p_t^A\left(x\right) -
  \nabla_x   \log p_t^B\left(x\right)  \right)\mathrm{d} x  - 0\\
\stackrel{}{=}& \mathbb{E}_{p_t^A} 
\left[ 
  u_t^A(x)  \cdot
  \left( \nabla_x  \log p_t^A\left(x\right) -
  \nabla_x   \log p_t^B\left(x\right) \right)
 \right]
\end{aligned}
\end{equation}

About the term \textcircled{2} in \Cref{eq:term1and2},
by using in sequence
$(i)$ 
Leibniz’s rule and Continuity Equation 
on $[\partial_t
 p_t^B\left(x\right)]$
, and $(ii)$
 the equality
$
\mathbf{\nabla} \cdot (\rho \mathbf{v})=\rho (\nabla \cdot \mathbf{v}) + (\mathbf{\nabla} \rho)  \cdot  v
$
if $\mathbf{v}$ is a vector field and $\rho$ is a scalar function,
we obtain
\begin{equation}\label{eq:term2}
    \begin{aligned}
 & \int_{\mathbb{R}^d}[
 p_t^A\left(x\right) ][\partial_t\log \left(\frac{p_t^A\left(x\right)}{p_t^B\left(x\right)}\right) ] \mathrm{d} x \\
=& \int_{\mathbb{R}^d}[
 p_t^A\left(x\right) ][
 \partial_t\log (p_t^A\left(x\right)) - \partial_t \log p_t^B\left(x\right)\ ] \mathrm{d} x   \\
 =& \int_{\mathbb{R}^d}[
 p_t^A\left(x\right) ][
 \frac{\partial_t p_t^A (x)}{p_t^A(x)} - \frac{\partial_t p_t^B (x)}{p_t^B(x)} ] \mathrm{d} x   \\
  =& \int_{\mathbb{R}^d}
[\partial_t p_t^A (x) -  p_t^A\left(x\right)  \frac{\partial_t p_t^B (x)}{p_t^B(x)} ] \mathrm{d} x   \\
  =& \int_{\mathbb{R}^d}
\partial_t p_t^A (x) \mathrm{d} x  -  \int_{\mathbb{R}^d} \frac{p_t^A\left(x\right) }{p_t^B(x)} \partial_t p_t^B (x)  \mathrm{d} x   \\
   \stackrel{(i)}{=}& \partial_t \int_{\mathbb{R}^d}
 p_t^A (x) \mathrm{d} x  -  \int_{\mathbb{R}^d} \frac{p_t^A\left(x\right) }{p_t^B(x)} (  -\nabla_x \cdot ( p_t^B(x) u_t^B(x)  ) ] \mathrm{d} x   \\
   \stackrel{(ii)}{=}& \partial_t 1  +  \int_{\mathbb{R}^d} \frac{p_t^A\left(x\right) }{p_t^B(x)} 
   \left(  
  u_t^B(x) \cdot \nabla_x p_t^B(x) +   p_t^B(x) \nabla_x  \cdot  u_t^B(x)  
  \right) 
  \mathrm{d} x   \\
  =& 0  +  \int_{\mathbb{R}^d} \frac{p_t^A\left(x\right) }{p_t^B(x)}  u_t^B(x)  \cdot \nabla_x  p_t^B(x) + \frac{p_t^A\left(x\right) }{p_t^B(x)}   p_t^B(x) \nabla_x  \cdot  u_t^B(x)     \mathrm{d} x   \\
=&  
\int_{\mathbb{R}^d} 
p_t^A\left(x\right) 
\left(
 \frac{ \nabla_x  p_t^B(x) }{p_t^B(x)}   \cdot u_t^B(x)  +
\nabla_x   \cdot u_t^B(x)    
\right)
\mathrm{d} x   \\
=&  
\int_{\mathbb{R}^d} 
p_t^A\left(x\right) 
\left(
\left(\nabla_x   \log p_t^B(x) \right) 
 \cdot u_t^B(x)  +
\nabla_x   \cdot u_t^B(x)    
\right)
\mathrm{d} x   \\
  \stackrel{(iii)}{=}& 0
    \end{aligned}
\end{equation}
in which the last equality $(iii)$ follows from Instantaneous Change of Variables:
recall that in Instantaneous Change of Variables (\Cref{eq:Instantaneous})
$\frac{\mathrm{d}}{\mathrm{~d} t} \log p_t\left(x\right)=-\operatorname{div}\left(u_t\right)\left(x\right)$,
its RHS can be rewritten as 
$ 
-\operatorname{div}\left(u_t\right)\left(x\right)
= 
- \nabla_{x}  \cdot  u_t(x)  
$;
using the chain rule, its LHS is equivalent to
$ 
\frac{\mathrm{d}}{\mathrm{~d} t} \log p_t\left(x\right)
= 
\frac{d \log p_t(x)}{d x }\ \cdot \frac{d x}{d t} 
= \left(\nabla_x \log p_t(x)\right) \cdot u_t(x)
$.
It follows that the Instantaneous Change of Variables yields $ \left( \nabla_x   \log p_t^B(x)  \right) \cdot u_t^B(x)   
 +
 \nabla_x  \cdot  u_t^B(x)   = 0$.

Finally, $(i)$ injecting \Cref{eq:term1} and \Cref{eq:term2} back into \Cref{eq:term1and2}, and $(ii)$ expressing the score functions with velocity fields (\Cref{eq:score2})  give: 
\begin{equation} \label{eq:term1and2_new}
\begin{aligned}
& 
\int_{\mathbb{R}^d}  
    p_{1}^A(x)
    \log \left(\frac{
    p_{1}^A(x)
    }{
   p_{1}^B(x)
    }\right) \mathrm{d} x
 \\
  \stackrel{(i)}{=}&  
\int_0^1 
\left[
\mathbb{E}_{p_t^A} 
\left[ 
  u_t^A(x)  \cdot
  \left( \nabla_x  \log p_t^A\left(x\right) -
  \nabla_x   \log p_t^B\left(x\right) \right)
 \right] 
 \right]  dt
 \\
  \stackrel{(ii)}{=}&  \int_0^1 
\mathbb{E}_{p_t^A} 
\left[ 
u_t^A(x)  \cdot
  \left[ \frac{t u_t^A(x) - x} {1-t} -
  \frac{t u_t^B(x) - x} {1-t} \right]
 \right] 
dt  \\
=& 
\int_0^1 
\mathbb{E}_{p_t^A} 
 \left[   
   \frac{t}{1-t}
u_t^A(x)  \cdot
  \left( u_t^A(x) - u_t^B(x) \right)
  \right]
dt
\end{aligned}
\end{equation}
i.e. \Cref{eq:to_prove} is proven. 
$\quad \quad \square$





\clearpage
\subsubsection{Proof of \Cref{sec:lemma2}}
\label{sec:proof_lemma2}

\begin{proposition}[Non-uniform sampling for importance sampling]
{
The inverse CDF of a PDF proportional to truncated $\frac{t}{1-t}$ is 
    \begin{equation}
        F_\epsilon^{-1}(u)= 
        \begin{cases}
            1 + W(-e^{-Zu-1}) & u \in\left[0, \frac{-\ln (1-t_\epsilon)-t_\epsilon }{Z}\right), 
        \\ 
            1 + \frac{1-t_\epsilon}{t_\epsilon} \left[\ln (1-t_\epsilon) + Zu\right] & u \in\left[\frac{-\ln (1-t_\epsilon)-t_\epsilon }{Z}, 1\right],
        \end{cases}
    \end{equation}
in which $W$ is the Lambert's $W$-function, and the normalizing constant is $Z=-\ln (1-t_\epsilon)$. 
}    
\end{proposition}


\textit{Proof.} 
Our goal is to sample from a PDF proportional to $\frac{t}{1-t}$ for most of its support. 
For some large $t_\epsilon \in [0,1]$, we define the following un-normalized density (\Cref{eq:truncated_pdf})
\begin{equation}
\label{eq:unnorm_pdf}
\tilde{f}_\epsilon(t)= \begin{cases}
\frac{t}{1-t} 
& t \in\left[0, t_\epsilon\right) \\ 
\frac{t_\epsilon}{1-t_\epsilon} 
& t \in\left[t_\epsilon, 1\right]
\end{cases}
\end{equation}
By integrating \Cref{eq:unnorm_pdf} w.r.t. $t$, 
we get the cumulative function of the unnormalzied density
\begin{equation}
\label{eq:unnorm_cdf}
\tilde{F}_\epsilon(t)= \begin{cases}
-\ln (1-t)-t 
& t \in\left[0, t_\epsilon\right) \\ 
-\ln (1-t_\epsilon)
+ \frac{t_\epsilon}{1-t_\epsilon} (t-1)
& t \in\left[t_\epsilon, 1\right]
\end{cases}
\end{equation}
Evaluating it at $t=1$ gives us the normalizing constant $Z=\tilde{F}_\epsilon(t=1)=-\ln (1-t_\epsilon)$, from which we obtain 
the CDF $F_\epsilon(t)=\frac{\tilde{F}_\epsilon(t)}{Z}$, 
and 
the inverse CDF that we need for sampling (using the inverse CDF transform)
\begin{equation} \label{eq:inverse_cdf}
F_\epsilon^{-1}(u)= 
\begin{cases}
1 + W(-e^{-Zu-1})
& u \in\left[0, 
\frac{
-\ln (1-t_\epsilon)-t_\epsilon 
}{Z}
\right) 
\\ 
1 + \frac{1-t_\epsilon}{t_\epsilon} \left[\ln (1-t_\epsilon) + Zu\right] 
& u \in\left[
\frac{
-\ln (1-t_\epsilon)-t_\epsilon 
}{Z}, 
1
\right]
\end{cases}
\end{equation}
in which $W$ is the Lambert's W-function.

Specifically, to solve the equation 
\begin{equation}
\label{eq:inverse_cdf_difficult}
    \frac{-\ln (1-t)-t}{Z} = u
\end{equation}
we denote $w:= 1-t$ and $b:= -Zu-1 $ for simplicity, then \Cref{eq:inverse_cdf_difficult} becomes 
$\ln w = b + w$, which could be rewritten as 
$-w e^{-w} = -e^b $,
therefore $-w$ is the Lambert W function $W(-e^b)$.
We note that the Lambert W function can be solved because 
$Z > 0$ and $u \geq 0$ give $-e^b \geq -e^{-1}$.
Furthermore, since $-e^b <0$, 
both the $W_0$ and $W_{-1}$ branches of the Lambert  W function are defined;
but since we are interested in the solution that remains in the range $-1 \leq W(x)$ to make \Cref{eq:inverse_cdf} well defined,
$W_0$ is the branch of interest.
$\quad \quad \square$


%% file: sections/experiments_hyperp.tex
\clearpage
\section{Experimental protocol details}\label{sec:finetuning_hyperparam}


We report in \Cref{tab:hyperparameters} all the hyper-parameters we used for our experiments.

\input{tables/hyperparameters}




To construct a fine-tuning set $\mathcal{S}$ based on point-wise MI, 
we use the pre-trained SD3.5-M and, given a prompt, conditionally generate 50 images with $\text{CFG}=4.5$ at resolution $1024 \times 1024$, while at the same time computing point-wise MI between the prompt and each image latent. Only the image with the highest MI is kept. This process is done twice for all the $700$ fine-tuning prompts $\mathcal{Y}$ defined by T2I-Combench. 
Given the constructed fine-tuning set, we finetune SD3.5-M for $2000$ iterations with LoRA adaptation.


Note that ($i$) there is no overhead at image generation time: once the pre-trained SD3.5-M has been fine-tuned with RFMI FT, conditional sampling takes the same amount of time of ``vanilla'' SD3.5-M and ($ii$) 
the time to process the workloads scales down (almost linearly) with the number of GPUs used according to our observations.

%% file: tables/hyperparameters.tex
\begin{table}[H]
\small
\centering
\caption{Training hyperparameters.}
\fontsize{7}{7}\selectfont
\label{tab:hyperparameters}
\begin{tabular}{
    @{$\:\:\:\:\:$} 
    l
    @{$\:\:\:\:\:$}
    r
}
\toprule
{\bf Name}  & {\bf Value}  \\ 
\midrule
Trainable model & MMDiT \\
\midrule
PEFT        & LoRA \\ 
Rank        &    $32$     \\
$\alpha$    &    $32$       \\  
\midrule
Learning rate (LR)  &  $5e-6$      \\ 
Gradient norm clipping    &     $0.005$    \\
\midrule
LR scheduler   &    Constant      \\
LR warmup steps  &      $400$   \\
\midrule
Optimizer  &      AdamW  \\
AdamW - $\beta_1$   &  $0.9$      \\
AdamW - $\beta_2$   &     $0.999$        \\
AdamW - weight decay & $1e-4$   \\
AdamW - $\epsilon$  & $1e-8$   \\
\midrule
Resolution & $1024 \times 1024$         \\
CFG scale &  4.5         \\
Denoising steps & $100$ \\
$M$ & 50 \\
$k$ & 1 \\
\midrule
Global batch size  & 240 \\
Training iterations & 2000 \\
\bottomrule
\end{tabular}
\end{table}